%% file: arxiv_main.tex
\documentclass{article}

\usepackage[colorlinks=true,linkcolor=blue,citecolor=blue,urlcolor=blue]{hyperref}
\usepackage[preprint]{neurips_2026}

\usepackage[utf8]{inputenc} 
\usepackage[T1]{fontenc}    
\usepackage{url}            
\usepackage{booktabs}       
\usepackage{amsfonts}       
\usepackage{nicefrac}       
\usepackage{microtype}      
\usepackage{xcolor}         
\usepackage{todonotes}
\usepackage{amsmath,amssymb, amsthm}

\usepackage{cleveref}
\usepackage{makecell}
\usepackage[labelfont=bf]{caption}
\input{def_neurips}

\title{Discrete Flow Matching: Convergence Guarantees Under Minimal Assumptions}

\author{%
  Le-Tuyet-Nhi Pham \\
     \'Ecole Polytechnique, CMAP, IP Paris \\
  Route de Saclay, 91120 Palaiseau, France\\
  \texttt{le-tuyet-nhi.pham@polytechnique.edu} \\
  \And
 Giovanni Conforti \\
    Università degli Studi di Padova,\\
 Via Trieste, 63, 35131 Padova, Italia \\
\texttt{gconfort@math.unipd.it} \\
 \And
 Zhenjie Ren \\
    LaMME, Universit\'e \'Evry Paris-Saclay  \\ Boulevard François Mitterrand
91000, France\\
  \texttt{zhenjie.ren@univ-evry.fr} \\
  \And
   Alain Durmus \\
     \'Ecole Polytechnique, CMAP, IP Paris \\
  Route de Saclay, 91120 Palaiseau, France\\
  \texttt{alain.durmus@polytechnique.edu}
}

\begin{document}

\maketitle

\begin{abstract}
Flow Matching has recently emerged as a popular class of generative models for simulating a target distribution $\mu_1$ from samples drawn from a source distribution $\mu_0$. This framework relies on a fixed coupling between $\mu_0$ and $\mu_1$, and on a deterministic or stochastic bridge to define an interpolating process between the two distributions. The time marginals of this process can then be approximately sampled by estimating the transition rates, or more generally the generator, of its Markovian projection.
This framework has recently been extended to the case of discrete source and target distributions, under the name \emph{Discrete Flow Matching} (DFM). However, theoretical guarantees for such models remain scarce. In this paper, we study two DFM models on $\mathbb{Z}_m^d = \{0,\ldots,m-1\}^d$, sampled through time discretization, and derive non-asymptotic associated bounds for both of them.
In contrast to previous work, we establish non-asymptotic bounds in Kullback--Leibler divergence for the early-stopped version of the target distribution. We also derive explicit convergence guarantees in total variation distance with respect to the true target distribution. Importantly, these bounds rely only on an approximation error assumption, relaxing standard score assumptions used in earlier works, while also yielding improved dependence on the vocabulary size $m$ and the dimension $d$.
\end{abstract}

\section{Introduction}
Generative modeling via stochastic processes has become a central paradigm in modern machine learning as illustrated by the popularity and interest in diffusion models \citep{sohl2015deep,song2019generative, ho2020denoising,song2021scorebased}. Recently, Flow Matching \citep{lipman2023flow, liu2022flow,albergo2022building, albergo2025stochastic}
has emerged as an alternative approach that defines a probability path between two arbitrary distributions and learns the characterization of a Markov dynamic that follows this path, resulting in simple training objectives and the ability to handle a wide variety of tasks.

More precisely, Flow Matching methods consist in simulating a target distribution $\mu_1$ from samples drawn from a source distribution $\mu_0$ over a finite time horizon. By employing a fixed coupling $\pi$ and a reference Markov process  $(R_t)_{t\in[0,1]}$, we can define an interpolant $(\iX_t)_{t\in [0,1]}$ such that $(\iX_0, \iX_1) \sim \pi$ and $(\iX_t)_{t\in [0,1]}$ given $(\iX_0,\iX_1)$ has the same conditional distribution as $(R_t)_{t\in [0,1]}$ given $(R_0,R_1)$.
However, $(\iX_t)_{t\in [0,1]}$ is not Markov in general.
One option then is to consider its Markov projection. Indeed, under appropriate conditions on the reference process, there exists a Markov process whose marginals are the same as the interpolated process. This process is called the Markov projection of $(\iX_t)_{t\in [0,1]}$. In particular, in the continuous setting, $(R_t)_{t\in [0,1]}$ is typically a diffusion process with constant volatility, possibly zero; hence encompassing deterministic flows. Then, the resulting Markov projection has the same volatility and has an unknown drift. However, this drift is a solution of a simple regression problem and can be approximated efficiently with additional conditions on $(R_t)_{t\in [0,1]}$ \citep{peluchetti2023non,albergo2022building, albergo2025stochastic,    lipman2023flow, liu2022rectified, liu2022flow}. The same type of construction can be adapted to the discrete data. In this setting, $(R_t)_{t\in [0,1]}$ is supposed to be a continuous-time Markov chain (CTMC) and a Markov projection of $(\iX_t)_{t\in [0,1]}$ also exists. This Markov process is also a CTMC whose generator or equivalently (concrete) score is a solution of a variational problem which can be efficiently solved under appropriate conditions on $(R_t)_{t\in [0,1]}$ \citep{lou2024discretediffusionmodelingestimating,holderrieth2024generator, gat2024discrete,campbell2024generative, shaul2024flow,wang2025fudoki}.
 This approach has shown strong empirical performance, mirroring the impact of discrete diffusion \citep{austin2021structured, hoogeboom2021argmax} and score-based methods \citep{campbell2022a, sun2023scorebased}, which are now supported by many theoretical guarantees \citep{pham2025discrete, bach2025sampling, conforti2025nonasymptoticconvergencediscretediffusion, liang2026sharpconvergenceratesmasked}.

Due to their increasing popularity, Flow Matching methods have motivated significant recent interest in understanding their theoretical foundations. While there has been a series of work providing such guarantees, e.g., \citep{silveri2024theoretical,liu2025finitetime}, in the continuous setting, achieving similar results as for continuous diffusion models,
such analyses are scarcer in the discrete setting. In particular, existing results often rely on either restrictive assumptions or uniformization scheme for simulating the associated CTMC \citep{ su2025theoreticalanalysisdiscreteflow, wan2025error}, or faces scalability issues as the vocabulary size $m$ increases \citep{wan2026corrected}.

	\paragraph{Our contribution.}

    This work addresses these existing limitations by investigating Discrete Flow Matching (DFM)  associated with two reference Markov processes: nearest-neighbor and uniform random walks. Leveraging the associated interpolants and supposing an approximation error on the estimated score, commonly assumed in the literature, we derive explicit convergence bounds of the resulting generative models adopting a Discrete Markov Probabilistic Model (DMPM) discretization scheme \citep{pham2025discrete}. More specifically,
    \begin{itemize}
        \item We demonstrate that under \textit{only} an  approximation error assumption, our proposed discretization with an early-stopping rule achieves an $O(\varepsilon^2)$ error in KL divergence with respect to the early-stopped version of the target distribution with a total step complexity of $\tilde O(d^2 / \varepsilon^2)$ for uniform random walk and $\tilde O(d^2m^2/\varepsilon^2)$ for nearest-neighbor random walk, where the notation $\tilde O$ hides logarithmic factors of $d,m, \varepsilon$.
        \item In addition, we also derive explicit convergence guarantees in total variation (TV) distance with respect to the true target distribution, which is missing in prior studies.
    \end{itemize}
    Our results not only improve the error's dependence on the vocabulary size $m$ and the dimension $d$ but also eliminate the restrictive boundedness conditions on the approximate score found in prior work \citep{su2025theoreticalanalysisdiscreteflow, wan2025error, wan2026corrected}.
    We provide a detailed comparison of these theoretical distinctions in \Cref{sec:existing_work}.

\paragraph{Notation.}
Given a measurable space $(\mse, \mce)$, we denote by $\mcp(\mse)$ the set of probability measures of $\mse$ and by $2^{\mse}$ the power set of $\mse$. Let $\mu, \, \nu \in \mcp(\mse)$, a probability measure $\pi \in \mcp(\mse \times \mse)$ is called a coupling of $(\mu, \nu)$ if its marginals satisfy $\pi(\msa \times \mse) = \mu(\msa)$ and $\pi(\mse \times \msb) = \nu(\msb)$ for any $\msa, \, \msb \in \mce$. We denote by $\Pi(\mu, \nu)$ the set of all couplings of $(\mu, \nu)$. We define the KL divergence (also called relative entropy) of $\mu$ with respect to $\nu$ as $\KL(\mu|\nu)\eqdef \int \log(\rmd \mu/\rmd \nu) \rmd \mu$ if \(\mu\) is absolutely continuous with respect to $\nu$, and $\KL(\mu|\nu)=+\infty$ otherwise.
Given two real numbers $a,b \in \R$, we write $a \lesssim b$ (resp. $a \gtrsim b$) to mean $a \leq Cb$ (resp. $a \geq Cb$) for a universal constant $C>0$. Given $n \in \N^*$, we denote $[n]=\{ 1,\ldots, n\}$. For any finite set $\msa$, let $|\msa|$ denote its cardinality.

\section{Discrete Flow Matching}

\subsection{Discrete Stochastic Interpolant and their Markov projection}\label{sec:dfm}
We first formalize Discrete Flow Matching (DFM) by constructing Markov jump dynamics that interpolate between the two distributions, denoted by $\mu_0$ and $\mu_1$, over the state space $\zset^d_m = \{0,1,\ldots,m-1\}^d$.
Our construction is similar to the original one introduced in \citep{gat2024discrete}. Our presentation nevertheless differs a little taking the stochastic interpolant perspective \citep{albergo2025stochastic} instead of the flow matching one.
To this end, we first define an interpolant $(\iX_t)_{t \in [0,1]}$ as follows. We refer to \Cref{app:ctmc} for the formal definition of the notions we used in this section and a proper introduction to CTMC.

\paragraph{Jump Markov Bridge and Discrete Stochastic Interpolant.}
Consider first a rate matrix (or generator) $(t,x,y) \in [0,1] \times (\Z^d_m)^2 \mapsto q_t(x, y)$. We assume $(q_t)_{t \in [0,1]}$ is a stable, conservative, and non-explosive rate matrix \citep{feinberg2014solutions}.  This rate matrix defines the transition probabilities between states of the  CTMC $(X_t)_{t \in [0,1]}$ with transition kernel denoted for $s,t,x_t,x_s\in ([0,1])^2 \times (\Z^d_m)^2$ with $s<t$ by $p_{t|s}(x_t|x_s) = \PP(X_t=x_t|X_s=x_s)$ if $\P(X_s=x_s) \neq 0$ and $p_{t|s}(x_t|x_s) = \P(X_t=x_t)$ otherwise. Indeed, if $(q_t)_{t \in [0,1]}$ is homogeneous, \ie, $q_t=q$ for any $t \in [0,1]$, the transition probabilities of  $(X_t)_{t \in [0,1]}$ are given for $s,t,x_t,x_s \in ([0,1])^2 \times (\Z^d_m)^2$ with $s<t$ by  $p_{t|s}(x_t|x_s) = \exp((t-s)q(x_s,x_t))$. This is an easy consequence of the Forward or Backward Kolmogorov equations given in  \eqref{eq:foward_kolm} and \eqref{eq:backward_kolm} respectively in the appendix.
Based on this CTMC, we can consider the jump Markov bridge measure $\B^q$ corresponding to the conditional distribution of $(X_t)_{t\in [0,1]}$ given it starting and end points $(X_0,X_1)$ on the space of càdlàg paths $\mathbf{D}([0,1], \mathbb{Z}_m^d)$, \ie, $\B^q(\cdot|X_0,X_1)$ represents the distribution of $(X_t)_{t \in [0,1]}$ given $(X_0,X_1)$; see \Cref{app:ctmc} for a formal definition.
Given a coupling $\pi \in \Pi(\mu_0,\mu_1)$,
we can now define
an interpolated process $(\iX_t)_{t\in [0,1]}$ that bridges $\mu_0$ and $\mu_1$ as:
	\begin{align}\label{def:interpolant}
		(\iX_0,\iX_1) \sim \pi \eqsp, \quad (\iX_t)_{t\in [0,1]} \big| (\iX_0, \iX_1) \sim \B^q(\cdot| \iX_0, \iX_1) \eqsp.
	\end{align}
	Denote by  $(\pI_t)_{t\in [0,1]}$ the time marginal densities of $(\iX_t)_{t\in [0,1]}$, that is  $\P(\iX_t \in \msa)= \sum_{x\in \msa}\pI_t(x)$ for $ t \in [0,1]$ and $\msa \subset \Z^d_m$. As a straightforward consequence of the definition of $(\iX_t)_{t\in [0,1]}$, we have
	\begin{equation}\label{def:bridge_density}
		\pI_t(x) = \sum_{x_0, x_1 \in \Z^d_m} p_{t|0}(x|x_0)p_{1|t}(x_1|x) \tilde{\pi}(x_0, x_1) \eqsp,
	\end{equation}
	where
	\(
	\tilde{\pi}(x_0,x_1) \eqdef {\pi(x_0, x_1)}/{p_{1|0}(x_1|x_0)} \eqsp.
	\)

	We consider here two reference dynamics for our framework: the Nearest-Neighbor Random Walk (NNRW) and the Uniform Random Walk (URW).

		\paragraph{Nearest-Neighbor Random Walk.}
	The  generator associated with the NNRW is defined as:
	\begin{equation}\label{def:generator_nnrw}
		q(x,y) = \begin{cases}1/2 & \text{if } y = \sigma(x) \text{ with } \sigma \in \Mcn \eqsp, \\
			-d & \text{if } y = x \eqsp,\\
			\quad 0 & \text{otherwise} \eqsp,\end{cases}
	\end{equation}
	where $\Mcn = \{ \sigma^\ell_+, \sigma^\ell_- \}_{\ell=1}^d$ denotes the set of operators performing forward and backward jumps along the $\ell$-th coordinate of the torus $\mathbb{Z}_m^d$:
	\begin{equation*}
		\sigma^\ell_{\pm}(x) = x \pm e_\ell \pmod m \eqsp.
	\end{equation*}
	Note that the base CTMC $(X_t)_{t\in [0,1]}$ has the component $X_t^i| X_0^i \overset{\text{i.i.d.}}{\sim} p_{t|0}^{(1)}$ for all $i \in [d]$ and $t\in [0,1]$. More precisely,
	for any component $i \in [d]$ and $0 \leq s<s+t \leq 1$, the displacement is given by $X_{t+s}^i - X_s^i = N_t^{i,+} - N_t^{i,-} \pmod m$, where $N_t^{i,\pm} \sim \mathrm{Poisson}(t/2)$ are independent counters for forward and backward jumps and the net displacement $S_t^i = N_t^{i,+} - N_t^{i,-}$ follows a Skellam distribution $\mathrm{Skellam}(t/2, t/2)$ \citep{skellam1946frequency}. Thus the transition density of the full process is the product of marginals on the torus $\Z^d_m$:
	\begin{equation}\label{eq:transition_density_nnrw}
		p_{t+s|s}(y|x) = \prod_{i=1}^d \sum_{k \in \mathbb{Z}} \mathbb{P} (S_t^i = y^i - x^i + km ) \eqsp,
	\end{equation}
	where the Skellam probability is expressed via the modified Bessel function of the first kind $I_{|a|}$:
	\begin{equation*}
		\mathbb{P}(S_t^i = a) = \rme^{-t} I_{|a|} (t) \eqsp, \quad I_b(z) = \sum_{n=0}^{\infty} \frac{1}{n! (n+b)!}\left(\frac{z}{2} \right)^{2n+b} \quad \text{for $a\in \Z, \, b \in \N$} \eqsp.
	\end{equation*}
	The formal proof establishing the validity of \eqref{eq:transition_density_nnrw} is detailed in \Cref{sec:poisson_expression_nnrw}.

	\paragraph{Uniform Random Walk.} The generator associated with the URW is defined as
	%
    \begin{equation}\label{eq:forward_generator_rw}
		q(x,y) = \begin{cases}
			\qquad 1/m & \text{if } d_{\mrh}(x,y)=1 \eqsp,\\
			- d(m-1)/m & \text{if } y=x \eqsp, \\
			\quad \qquad 0 & \text{otherwise} \eqsp,
		\end{cases}
	\end{equation}
	where $d_{\mrh}(x,y)$ denotes the Hamming distance between $x$ and $y$. With a slight abuse of notation, we denote the set of possible jump operators as $\Mcu \eqdef \{ \sigma^\ell_n \, : \, \ell \in [d], \, n \in [m-1] \}$, where \[\sigma^\ell_n(x) = x + n e_\ell \pmod m\] describes a jump of size $n$ on the $\ell$-th component. Using the properties of the Hamming distance on product spaces, the transition density $p_{s+t|s}(y|x)$ can be derived \citep[Proposition 1]{zhang2024convergence}:
	\begin{equation}\label{eq:transition_density_urw}
		p_{s+t|s}(y|x) = \left( \frac{1+(m-1)\rme^{-t}}{m} \right)^{d} \alpha_t^{d_{\mrh}(y,x)} \eqsp, \quad \text{where }\quad  \alpha_t \eqdef \frac{1-\rme^{-t}}{1+(m-1)\rme^{-t}} \eqsp.
	\end{equation}

    \paragraph{Markovian projection of the interpolant.}
	A fundamental challenge is that the interpolant $(\iX_t)_{t\in [0,1]}$ is generally non-Markovian. To address this issue, we introduce its Markovian projection $(\mX_t)_{t\in [0,1]}$. As we see in our next result, this process is a Markov process uniquely characterized by a time-inhomogeneous generator that preserves the marginal distributions of the original path, that is $\P(\mX_t=x) = \P(\iX_t=x) = \pI_t(x)=:\mu_t(x)$ for all $t \in [0,1]$, $x \in \Z^d_m$, and therefore that transports the marginals $\mu_0$ to $\mu_1$.

	\begin{theorem}[Markovian Projection of the Mixture Bridge]\label{theo:markov_projection}
		Let $\pi \in \Pi(\mu_0, \mu_1)$ be an arbitrary coupling and $q$ be a homogeneous stable, conservative, and non-explosive generator. The Markovian projection $(\mX_t)_{t\in [0,1]}$ of $(\iX_t)_{t\in [0,1]}$ defined in \eqref{def:interpolant} is the unique (in law) Markov process characterized by the time-inhomogeneous generator $\mq_t$ such that, for all $t \in [0,1]$ and $x\neq y \in \Z^d_m$:
        \begin{equation}\label{def:markov_generator_u}
			\mq_t(x,y) = q(x,y) \uM_t(x,y) \eqsp,
		\end{equation}
		where the score function $\uM$ is defined as
		\begin{equation}\label{def:score}
			\uM_t(x,y) = \frac{\sum_{x_0,x_1\in \Z^d_m} p_{1|t}(x_1|y)p_{t|0}(x|x_0) \tilde\pi(x_0,x_1)}{\pI_t(x)}
            \eqsp.
		\end{equation}
	\end{theorem}
	For completeness, the full proof of \Cref{theo:markov_projection} is provided in \Cref{proof_theo:markov_projection}.
    Note that $\uM$ can be expressed as the conditional expectation
    \begin{align}\label{eq:score_cond_exp}
         \uM_t(x,y) = \E \l[ u_t^{(\iX_1)}(x,y) \middle| \iX_t =x\r] \quad \text{where } u_t^{(x_1)}(x,y) \eqdef \frac{p_{1|t}(x_1|y)}{p_{1|t}(x_1|x)} \eqsp.
    \end{align}
    This implies \begin{equation}\label{def:markov_generator}
			\mq_t(x,y) := \mathbb{E} \left[ q_t^{(\iX_1)}(x,y) \mid \iX_t = x \right] \quad \text{where} \quad q_t^{(x_1)}(x,y) = q(x,y) u_t^{(x_1)}(x,y)  \eqsp.
		\end{equation}

	\subsection{Generative Process}

	While the Markovian projection defines an ideal generative process, its exact simulation remains infeasible due to two main bottlenecks: (i) the intractability of the exact score defined in \eqref{def:score}, and (ii) the requirement for time discretization in inhomogeneous CTMCs. Although techniques such as the uniformization algorithm \citep{wan2025error} can bypass discretization, they can suffer from high computational cost and poor scalability in high-dimensional regimes. We address the former by learning an approximate score via the loss function detailed below, and the latter by employing DMPM discretization scheme \citep{pham2025discrete}, implementing a piecewise-constant generator. In the following sections, we formalize our generative framework, detailing the score training procedure and the construction of the resulting approximate dynamics.
	\paragraph{Practical Training and Loss Minimization.}
	To approximate the true discrete score $\uM$, we introduce a parameterized function class $\{\uMtheta : [0,1] \times (\mathbb{Z}^d_m)^2 \to \mathbb{R}_+ \, : \, \theta \in \Theta \}$. Given a time-discretization $\{t_k\}_{k=0}^K$  of $[0,1-\eta]$ associated with the step-size $h_{k+1} = t_{k+1} - t_k$ where $\eta \in (0,1)$ denotes an early-stopping parameter, the optimal parameter $\theta^\star$ is obtained by minimizing the following empirical divergence-based objective:
\begin{equation}\label{eq:loss_main}
		\theta \mapsto \loss_{\mre}(\theta) \eqdef \sum_{k=0}^{K-1} h_{k+1}  \E \l[ \sum_{\sigma \in \Mc} \mqtheta_{t_k} \varphi \l(\frac{\mq_{t_k}}{\mqtheta_{t_k}}\r)(\iX_{t_k}, \sigma(\iX_{t_k})) \r] \eqsp,
	\end{equation}
	where $\mq$ is defined in \eqref{def:markov_generator_u}, $\mqtheta = q\uMtheta$, $\varphi(a) = a\log(a) +1-a$ for $a\geq 0$, and $\Mc$ serves as a unified shorthand for both $\Mcn$ and $\Mcu$ for brevity. This is discrete version of the score entropy loss used in \cite{lou2024discretediffusionmodelingestimating,gat2024discrete}. It has been already considered in the literature on theoretical guarantee for discrete diffusion models \cite{pham2025discrete, liang2025discrete, liang2025absorb, dmitriev2026efficient}.
We also consider a discretized version of the concrete score matching loss considered in \cite{meng2022concrete}
    \begin{align*}
        \loss_2 (\theta) \eqdef \sum_{k=0}^{K-1} h_{k+ 1} \E \l[\sum_{\sigma \in \Mc} \l( \l( \mqtheta_{t_k} - \mq_{t_k}  \r) (\iX_{t_k}, \sigma(\iX_{t_k})) \r)^2 \r] \eqsp.
    \end{align*}

    The final objective on which we will make our main approximation assumption is obtained by combining the two previous losses:
    \begin{equation}
      \label{eq:def_loss_tot}
\loss = \loss_{\mre} + \loss_2 \eqsp.
    \end{equation}
    Following \cite{meng2022concrete,lou2024discretediffusionmodelingestimating}, $\loss$ admits the following numerically tractable simplification, thanks to the conditional expectation expression of the generator in \eqref{def:markov_generator}:
    \begin{align*}
        \tilde\loss (\theta)  =  \sum_{k=0}^{K-1} h_{k+1} \E \l[\sum_{\sigma \in \Mc} \l(-q_{t_k}^{(\iX_1)} \log \mqtheta_{t_k} + \mqtheta_{t_k} + \l( \mqtheta_{t_k} - q^{(\iX_1)}_{t_k} \r)^2  \r)(\iX_{t_k}, \sigma(\iX_{t_k}))    \r] \eqsp,
    \end{align*}
where $q_t^{(x_1)}(x, \sigma(x))$ admits the closed-form formula in \eqref{def:markov_generator}.

	%

	\paragraph{Generative Dynamics.}
	Given the learned score $\{\uMthetastar_{t_k}\}_{k=0}^K$ associated with the discretization grid $\{t_k\}_{k=0}^K$,
	we define the generative process as follows. Starting from the prior distribution $X^{\theta^\star}_0 \sim \mu_0$, we propagate the state forward using the approximate score. For each $t \in [t_k, t_{k+1})$, given $X^{\theta^\star}_{t_k}$, the dynamics are governed by the generator $q^{\mathrm{M},\theta^\star}_t(x,y) = \tuMthetastar_{t}(x,y| X^{\theta^\star}_{t_k}) q(x,y) $
	for $x \neq y$, where $\tuMthetastar$ is given by \( \tuMthetastar_t(x,y | X^{\theta^\star}_{t_k}) = \uMthetastar_{t_k}(X^{\theta^\star}_{t_k}, \sigma (X^{\theta^\star}_{t_k}))  \) if $y = \sigma (x)$ for $\sigma\in \Mc$ and is $0$ otherwise.
	Equipped with this generator, we can use an exponential distributed holding time to sample $(\Xthetastar_t)_{t\in [t_k,t_{k+1}]}$. This approach, referred to as DMPM, is adopted from \citep{pham2025discrete}. Crucially, we stop the simulation at $t_K = 1-\eta$ to ensure the convergence results hold despite the score's divergence at the final time step.

%
%
%

	\section{Main Results}\label{sec:convergence}

	\subsection{Convergence Guarantees of DFM}
	To quantify the performance of DFM, we first formalize the cumulative approximation error across the training intervals.
	\begin{assumption_main}[Approximation Error]\label{ass:approx_error_2}
		There exists a constant $\tilde{\varepsilon} \geq 0$ such that
		\[
\loss(\theta^{\star}) \leq \tilde{\varepsilon}^2 \eqsp.
		\]
	\end{assumption_main}
        
        Under this requirement, we establish our main result as follows.
	\begin{theorem}\label{theo:main_2}
		Consider the early-stopping parameter $\eta \in (0,1)$ and the time-discretization $\{t_k\}_{k=0}^K$ of $[0,1-\eta]$ with
			\(
		t_k = 1 - (1+h)^{-k}
		\), where $h \in (0,1)$ is the parameter that controls the step sizes. Under \Cref{ass:approx_error_2}, the following holds for both NNRW and URW
		\begin{align*}
			\KL(\mu_{1-\eta}| \mathrm{Law}(\Xthetastar_{1-\eta})) &\lesssim   m^2 \eta^{-2} \tilde{\varepsilon}(\sqrt{|\Mc|}+\tilde{\varepsilon}) + ({\lambda(m)})^2d^2 m^2 h \log(\eta^{-1})\log \l(m{\eta}^{-1} \r)  \eqsp,
		\end{align*}
		where $\lambda(m)$  represents the fixed jump rate, taking value $1/2$ in the case of NNRW and $1/m$ for URW, and $\Mc$ denotes the collection of jump operators  (see \Cref{sec:dfm} for detailed definition). 
	\end{theorem}
	\Cref{theo:main_2} derives a non-asymptotic error bound for DFM  under minimal assumption \Cref{ass:approx_error_2}, explicitly decomposing the KL divergence into approximation and discretization components. We provide a sketch of the proof for this main theorem in \Cref{sec:method}, while the complete technical derivation is deferred to \Cref{proof_theo:main_2}.

		\Cref{theo:main_2} indicates that to reach a precision of $\KL(\mu_{1-\eta}|\mathrm{Law}(\Xthetastar_{1-\eta})) = O(\varepsilon^2)$, the parameter controlling step size and the approximation error can be set as:
        \[h = \frac{\varepsilon^2 }{d^2m^2(\lambda(m))^2\log(\eta^{-1})\log(m\eta^{-1})} \quad \text{and} \quad  \tilde{\varepsilon}  = \frac{\varepsilon^2 \eta^2}{m^2 \sqrt{|\Mc|}} \eqsp.\]
        This leads to a total iteration complexity of:
		\[
		K = \frac{\log(\eta^{-1})}{\log(1+h)} = O \l(  \frac{d^2 m^2(\lambda(m))^2}{\varepsilon^2} \log^2\frac{1}{\eta} \log \frac{m}{\eta}  \r) \eqsp.
		\]
        This quadratic scaling with respect to the dimension $d$ is consistent with existing complexity findings in \citep[Corollary 5.2]{liu2025finitetime} and \citep{wan2026corrected}
        .
       In particular, we obtain an iteration complexity of $\tilde O(d^2/\varepsilon^2)$ for URW and $\tilde O(d^2m^2/\varepsilon^2)$ for NNRW setting (where $\tilde O$ omits logarithmic factors).
		 Crucially, for the URW case, our analysis  improves the dependence on the vocabulary size $m$ compared to the recent work of \citep{wan2026corrected}.
	\begin{corollary}\label{cor:tv_distance_2}
		Consider the early-stopping parameter $\eta \in (0,1)$ and the time-discretization $\{t_k\}_{k=0}^K$ of $[0,1-\eta]$ with
			\(
		t_k = 1 - (1+h)^{-k}
		\), where $h \in (0,1)$ is the parameter that controls the step sizes. Under \Cref{ass:approx_error_2}, the following holds for both NNRW and URW
		\begin{align*}
				\| \mathrm{Law}(\Xthetastar_{1-\eta}) - \mu_1 \|_{\TV} \lesssim {\mrc(m) d\eta} + m\eta^{-1}\sqrt{\sqrt{|\Mc|} \tilde{\varepsilon} + \tilde{\varepsilon}^2 }+ \lambda(m){d m\sqrt{h \log \l(m\eta^{-1} \r)\log(\eta^{-1}) }} ,
		\end{align*}
		where $\mrc(m)$is defined as $1$ for the URW and $m$ for the NNRW dynamics. Here $\lambda(m)$  represents the fixed jump rate, taking value $1/2$ in the case of NNRW and $1/m$ for URW, and $\Mc$ denotes the collection of jump operators 
	\end{corollary}
	\Cref{cor:tv_distance_2} establish the explicit error bound in TV distance with respect to the true target distribution, which is missing in \citep{wan2026corrected}. The proof
    follows directly from \Cref{theo:main_2} through an application of Pinsker’s and triangle inequalities. For completeness, the detailed proof of \Cref{cor:tv_distance_2} is provided in \Cref{proof_cor:tv_distance_2}.

	\begin{remark}
		The term $\mrc(m)d\eta$ quantifies the early-stopping error, a direct consequence of truncating the generative flow at $1-\eta$ to avoid the singularity of the score at $t=1$.
		 In the URW case, the graph is effectively complete, \ie, any state can transition to any other in a single jump. This all-to-all connectivity allows the distribution to shift globally and instantaneously, leading to a bound that is independent of the state-space size ($\mrc(m)=1$). In contrast, the NNRW is constrained to a lattice-like topology where transitions only occur between immediate neighbors. To move probability mass across the domain, the NNRW must flow through $m$ intermediate states, which is reflected in the linear scaling $\mrc(m)=m$. 
         \Cref{cor:tv_distance_2} implies that to achieve $O(\varepsilon)$ error in TV distance, we can set
\begin{align*}
    \eta = \frac{\varepsilon}{\mrc(m)d}\eqsp, \quad \tilde{\varepsilon}  = \frac{\varepsilon^4}{(\mrc(m))^2 d^2m^2 \sqrt{|\Mc|}}   \quad \text{and} \quad h = \frac{\varepsilon^2}{(\lambda(m))^2 d^2 m^2 \log(m\eta^{-1})\log(\eta^{-1})} \eqsp.
\end{align*}
This implies a total iteration complexity of:
\begin{align*}
    K = \frac{\log(\eta^{-1})}{\log(1+h)} = \tilde O\l(\frac{d^2 m^2 (\lambda(m))^2 }{\varepsilon^2} \r) \eqsp.
\end{align*}
In particular, we get an iteration complexity $ \tilde O(d^2/\varepsilon^2)$ for URW and $\tilde O(d^2 m^2/\varepsilon^2)$ for NNRW setting.
	\end{remark}

	\subsection{Proposed methodology: A Sketch of Proof}\label{sec:method}
	\paragraph{Convergence Proof.}
    In what follows, we present a sketch of proof of the main result stated in \Cref{theo:main_2} in order to outline our core methodology.
	To evaluate KL divergence of ground-truth against the generated distribution, we first employ the data processing inequality for the relative entropy \citep[Lemma 1.6]{nutz2021introduction}. We then decompose the total divergence into propagation mismatches by applying Girsanov's theorem for jump processes \citep[Theorem G.0.1]{conforti2025nonasymptoticconvergencediscretediffusion} and substitute the formula of the piecewise-constant generator as follows
	\begin{align*}
	 \KL(\mu_{1-\eta}|\mathrm{Law}&(\Xthetastar_{1-\eta}))\\
	 &\leq  \KL(\mathrm{Law}(\mX_{[0,1-\eta]}) | \mathrm{Law}(\Xthetastar_{[0,1-\eta]})) \notag \\
		&= \lambda(m) \sum_{k=0}^{K-1} \int_{t_k}^{t_{k+1}}  \E \Bigg[ \sum_{\sigma \in \Mc}  \l( {\uM_t}\log \frac{\uM_t}{\uMthetastar_{t_k}} - {\uM_t}+\uMthetastar_{t_k} \r)  (\sigma) \Bigg] \rmd t \notag\\
		\leq & \lambda(m) \underbrace{\sum_{k=0}^{K-1} \int_{t_k}^{t_{k+1}}   \E \l[ \sum_{\sigma \in \Mc}\l( \varphi ( \uM_t)  - \varphi (\uM_{t_{k}})+ \log \uM_{t_k} (\uM_{t_k} - \uM_t) \r) (\sigma)\r]  \rmd t}_{E_{\text{disc}}}\\
        &\qquad + \underbrace{\sum_{k=0}^{K-1} \int_{t_k}^{t_{k+1}}   \E \l[ \sum_{\sigma \in \Mc}\l(\mq_t\log \frac{\mq_{t_k}}{\mqthetastar_{t_k}} - \mq_{t_k} + \mqthetastar_{t_k} \r) (\sigma)\r]  \rmd t}_{E_{\text{approx}}}  \eqsp,
	\end{align*}
	where $\varphi(a) \eqdef a\log (a) -a+1$ and $\lambda(m)$ represents the fixed jump rate, taking value $1/2$ for NNRW and $1/m$ for URW. Here we used the shorthand $\mq_t(\sigma) = \mq_t(\mX_t, \sigma(\mX_t))$ (and analogously for $\mq_{t_k}(\sigma), \mqthetastar_{t_k}(\sigma)$). To analyze the approximation error $E_{\text{approx}}$, we decompose the rate $\mq_t$ as $(\mq_t/\mq_{t_k} )\mq_{t_k}$ to isolate the divergence term $\mq_{t_k}\log(\mq_{t_k}/\mqthetastar_{t_k})$. We then use the following boundedness of the score to upper bound the ratio $\mq_t/\mq_{t_k}$ by $m^2/\eta^2$:
    \begin{lemma}[Uniform Boundedness of the Score]\label{prop:bound_score}
		For both the NNRW and URW base processes, the score is uniformly bounded on $[0, 1-\eta]$. Specifically, for any $x \in \Z^d_m, \, t \in  [0,1-\eta]$ and $\sigma \in \Mc$,
		\begin{equation*}
			\frac{1-t}{m}  \lesssim \uM_t (x, \sigma(x)) \lesssim \frac{m }{1-t}   \eqsp.
		\end{equation*}
        Additionally, for any $x \in \Z^d_m$ and $t\in [0,1-\eta]$, the following bounds hold for any $x_1 \in \Z^d_m$:
        \begin{equation}\label{eq:bound_sum_score}
        \sum_{\sigma \in \Mc} u_t^{(x_1)} (x,\sigma(x)) \lesssim \frac{dm}{1-t} \quad \text{and} \quad
            \sum_{\sigma \in \Mc} \uM_t(x,\sigma(x)) \lesssim \frac{dm}{1-t} \eqsp.
        \end{equation}
	\end{lemma}
    Consequently, bounding $E_{\text{approx}}$ reduces to controlling the term $\mq_{t_k}\log(\mq_{t_k}/\mqthetastar_{t_k})$ and $\mqthetastar_{t_k} - \mq_{t_k}$,
    which can be addressed by applying triangle, Cauchy-Schwarz and Jensen inequalities, then using 
    \[
    \sum_{k=0}^{K-1} h_{k+1}  \E \l[\sum_{\sigma \in \Mc}\l\{ \mqthetastar_{t_k} \varphi \l(\frac{\mq_{t_k}}{\mqthetastar_{t_k}}\r) + \l( \mqthetastar_{t_k} - \mq_{t_k} \r)^2 \r\} (\mX_{t_k}, \sigma(\mX_{t_k}) ) \r] \leq \tilde{\varepsilon}^2 \eqsp, 
    \]
    which is equivalent to \Cref{ass:approx_error_2} since $\mX_t \overset{\text{dist}}{=} \iX_t$ for all $t \in [0,1]$. We then obtain $ E_{\text{approx}} = O(m^2 \eta^{-2} \tilde{\varepsilon}(\sqrt{|\Mc|} +\tilde{\varepsilon} ))$.
    To analyze $E_{\text{disc}}$, we first derive an ordinary differential equation  (ODE) for the DFM score by leveraging the Kolmogorov equations \eqref{eq:foward_kolm} \eqref{eq:backward_kolm}:
	\begin{lemma}\label{prop:hjb}
		For any fixed $\sigma \in \mathcal{M}$, the score satisfies the following ODE
    \begin{equation}\label{eq:hjb_u}
			\begin{aligned}
				\partial_t \uM_t(x,\sigma(x)) =  \lambda(m)&\sum_{\sigma' \in \Mc} \mcam_t(x, \uM; \sigma, \sigma')
				\eqsp,
			\end{aligned}
		\end{equation}
		where $\lambda(m)$ represents the fixed jump rate, taking the value $1/2$ in the case of NNRW and $1/m$ for URW.  Here the operator $\mcam_t$ is defined by
		\begin{align*}
			\mcam_t(x,\uM; \sigma, \sigma') \eqdef
			&\uM_t(x,\sigma(x)) \uM_t(x, \sigma' (x))-\uM_t(x,\sigma' (\sigma(x))) \\
			&+ \frac{\pI_t(\sigma' (x))}{\pI_t(x)} \l[ \uM_t(\sigma' (x), \sigma(x)) - \uM_t(x,\sigma(x)) \uM_t(\sigma' (x),x) \r] \eqsp.
		\end{align*}
	\end{lemma}
    This ODE allows us to obtain explicit evolutions for the score and its entropy function by applying It\^o's formula:

	\begin{lemma}[Score and Entropy Dynamics]\label{prop:evolution_score}
		Fix $\sigma \in \Mc$, the evolution of the score $\uM_t(\sigma)$ and its associated entropy $\varphi(\uM_t(\sigma))$ are governed by:
		\begin{equation}\label{eq:evolution_u}
			\begin{aligned}
				 \E \l[\uM_t(\sigma) \middle| \Fc_s \r] - \uM_s(\sigma) &= \lambda(m) \int_s^t  \E  \l[\sum_{\sigma' \in \Mc} \mcbm_r(\mX_r, \uM; \sigma, \sigma') \middle| \Fc_s \r] \rmd r \eqsp, \\
                 \E\l[ \varphi(\uM_t(\sigma))\r] - \E \l[\varphi(\uM_s(\sigma)) \r] &= \lambda(m) \int_s^t  \E \l[\sum_{\sigma' \in \Mc} \mccm_r(\mX_r, \uM;\sigma, \sigma') \r] \rmd r \eqsp,
			\end{aligned}
		\end{equation}
		for any $0\leq s \leq t \leq 1-\eta$, where $(\Fc_t)_{t\in [0,1-\eta]}$ denotes the filtration of $(\mX_t)_{t\in [0,1-\eta]}$ and the operators $\mcbm, \mccm$ are defined by
		\begin{align*}
			\mcbm_r  (x, \uM; \sigma, \sigma') \eqdef  & \uM_r(x,\sigma' (x)) \uM_r(\sigma' (x),\sigma( \sigma' (x)))  -\uM_r(x,\sigma' (\sigma(x)))\\
			&\qquad+ \frac{\pI_r(\sigma' (x))}{\pI_r(x)} \l[ \uM_r(\sigma' (x), \sigma(x)) - \uM_r(x,\sigma(x)) \uM_r(\sigma' (x),x) \r] \eqsp.
            \\
        \mccm_r(x, \uM; \sigma, \sigma') \eqdef &\uM_r(x, \sigma(\sigma' (x))) \log \frac{\uM_r(\sigma' (x),\sigma(\sigma' (x)))}{\uM_r (x, \sigma(x))} \\
			&\qquad + \uM_r (x, \sigma' (x)) \l[ \uM_r(x, \sigma(x)) - \uM_r(\sigma' (x), \sigma(\sigma' (x))) \r] \eqsp.
		\end{align*}
	\end{lemma}
    Substituting these expressions into discretization error gives
	\begin{align*}
& 		 E_{\text{disc}}
           \\
          &\leq \lambda(m)\sum_{k=0}^{K-1}\int_{t_k}^{t_{k+1}} \int_{t_k}^t \E \l[\sum_{\sigma, \sigma' }  \l(\mccm_r(\mX_r, \uM; \sigma, \sigma') - \log \uM_{t_k}(\sigma) \mcbm_r(\mX_r, \uM; \sigma, \sigma') \r)\r] \rmd r \rmd t \eqsp.
	\end{align*}
    The primary technical challenge in the above expression lies in logarithmic terms. If the score vanishes or explodes near the boundaries of the time interval, the discretization gap becomes uncontrollable. Our analysis hinges on a key stability property shown in \Cref{prop:bound_score} that ensures the discrete score remains bounded on $[0,1-\eta]$.
    This boundedness ensures that the logarithmic term is stable up to the early-stopping threshold $\eta$. By strategically selecting the time-grid $\{t_k\}_{k=0}^K$ to counteract the $1/(1-t)$ growth, we close the estimates and achieve the $\tilde{O}((\lambda(m))^2d^2m^2h)$ total discretization error in KL divergence.

	\begin{remark}
		In standard discrete diffusion models, the score function satisfies a transitive identity: $u_t(x, z) = u_t(x, y)u_t(y, z)$, which causes the integrand in \eqref{eq:evolution_u} to vanish, greatly simplifying the convergence analysis. In contrast, DFM lacks this property, rendering its theoretical analysis  more complex than its diffusion-based counterparts.
	\end{remark}

	\paragraph{Proof of Stability of the Score.}
	The proof of the key Lemma \ref{prop:bound_score} relies on the characterization of the projected score $\uM_t$ as the conditional expectation of the bridge score $u_t^{(x_1)}$:
	\begin{align*}
		\uM_t (x, y) = \E \l[ u_t^{(\iX_1)}(x, y) \middle| \iX_t =x \r]  \eqsp,
	\end{align*}
	for $x,y \in \Z^d_m$ and $t \in [0,1-\eta]$. Here, $u_t^{(x_1)}$ denotes the bridge score, which is given by the ratio of transition densities:
	\begin{align*}
		u_t^{(x_1)}(x, y) = \frac{p_{1|t}(x_1|y)}{p_{1|t}(x_1|x)} \eqsp.
	\end{align*}
	Consequently, the projected score inherits the uniform stability of the underlying bridge score. For the NNRW, the boundedness of $u_t^{(x_1)}$ is established by analyzing the ratios of modified Bessel functions within the Skellam density, leveraging the Amos inequality \citep[Eq. (9)]{amos1974computation} to bound these ratios. For the URW, stability follows directly from the explicit transition density formula, which ensures the discrete score remains bounded for all $t \in [0, 1-\eta]$.
	We emphasize that an \textit{early-stopping strategy} is essential here, as the score exhibits a singularity at the terminal time $t=1$. The detailed proof is provided in \Cref{proof_prop:bound_score} for completeness.

%

	\section{Related works}\label{sec:existing_work}
    \paragraph{Discrete Diffusion Models.}
    Discrete diffusion models (DDMs)  are central to categorical generation in domains like language and chemistry \citep{austin2021structured, hoogeboom2021argmax}. Recent works treats their reverse dynamics as controlled jump processes to derive non-asymptotic bounds \citep{pham2025discrete, conforti2025nonasymptoticconvergencediscretediffusion, liang2025discrete,liang2026sharp}.


    \paragraph{Discrete Flow Matching.}
    Despite DFM’s practical success \citep{gat2024discrete, campbell2024generative}, theoretical research remains limited. Existing analyses \citep{su2025theoreticalanalysisdiscreteflow, wan2025error} typically focus on analyzing the statistical error of the one-jump-bridge-based dynamic using the exact uniformization simulation.
    While such algorithm eliminates discretization error theoretically, it can suffer from poor scalability. In contrast, our work implements the DMPM sampler \citep{pham2025discrete} using the random-walk-based dynamics  and focus on analyzing the resulting {discretization error}.
    Unlike prior studies \citep{su2025theoreticalanalysisdiscreteflow, wan2025error}, our framework requires no assumptions on the ground-truth or estimated generators. Furthermore, we achieve a  reduction in error dependence on both the dimension $d$ and vocabulary size $m$.
    In comparison to the recent work of \citep{wan2026corrected}, our analysis provides a tighter complexity bound with respect to $m$ by leveraging the favorable properties of URW dynamic. In addition, it is worth mentioning that the assumption of the approximation error considered in \citep{wan2026corrected} is different from ours. Indeed, in our notation, their condition reads:
    \begin{equation}
      \label{eq:assum_approx_other}
      \sum_{k=0}^{K-1} \E\l[ \int_{t_k}^{t_{k+1}}  \E \l[ \sum_{z \neq \Xalg_t} \mqthetastar_{t} \varphi \l(\frac{\qalg_{t}}{\mqthetastar_{t}}\r)(\Xalg_{t}, z) \middle| \Xalg_{t_k} \r] \rmd t \r] \leq \tilde{\varepsilon}^2 \eqsp,
    \end{equation}
    where $(\Xalg_t)_{t\in [0,1]}$ is a process with the transition rate $(\qalg_t)_{t\in [0,1]}$ induced by the ground-truth generator and the specific time-discretization scheme. 
    We summarize this comparison in \Cref{tab:comparison}.   

\begin{table}[ht]
        \centering
        \footnotesize
        \begin{tabular}{|c|c|c|c|c|c|}
        \hline
          \textbf{ References}   & \textbf{Discretization scheme} &\makecell{\textbf{Approx.} \\ \textbf{assump.}} & \makecell{\textbf{Early-}\\ \textbf{stopping*}} &  \textbf{Measure} & \textbf{Iteration Complexity}  \\
          \hline
             \makecell{\citep{wan2026corrected} \\ Theorem 1}&{Tau-leaping and Euler } & \eqref{eq:assum_approx_other}&Yes
             & TV distance & $\tilde O(d^2 m /\varepsilon^2)$ \\
             \hline
             \makecell{\citep{wan2026corrected} \\ Theorem 3} & {Time-corrected sampler
             } & \eqref{eq:assum_approx_other}& Yes & TV distance &$\tilde O(d^2m/\varepsilon^2)$ \\
             \hline
             \makecell{\citep{wan2026corrected} \\
             Theorem 4} & Location-corrected sampler &\eqref{eq:assum_approx_other} & Yes & TV distance & \makecell{$\tilde O(d^2m/\varepsilon^2)$ \\(worst case)} \\
             \hline
             \makecell{\textbf{This work} \\
            \textbf{\Cref{theo:main_2}}} & \textbf{DMPM with URW dynamic} &\Cref{ass:approx_error_2}& \textbf{Yes} & \textbf{KL divergence} & $\tilde O(d^2/\varepsilon^2)$ \\
             \hline
             \makecell{\textbf{This work} \\
            \textbf{\Cref{cor:tv_distance_2}}} & \textbf{DMPM with URW dynamic} &\Cref{ass:approx_error_2} & \textbf{No} & \textbf{TV distance} & $\tilde O(d^2/\varepsilon^2)$ \\
             \hline
        \end{tabular}
        \medskip
        \caption{Comparison of complexity results for achieving $O(\varepsilon)$ error in TV distance or $O(\varepsilon^2)$ error in KL divergence between our work and existing theoretical DFM literature. \textit{Early-stopping:} ``Yes'' denotes that the error bound applies to an early-stopped version of the target distribution, whereas ``No'' signifies a bound derived for the true target distribution. }
        \label{tab:comparison}
    \end{table}

\section{Conclusion and Limitations}\label{sec:conclusion}
We establish the non-asymptotic convergence guarantees for DFM on $\mathbb{Z}_m^d$ that rigorously account for discretization error under minimal assumptions. By characterizing the score dynamics through our framework and leveraging score's stability, we address the technical challenges posed by DFM's.
We first establish convergence guarantees in KL divergence for DFM relative to the early-stopped target distribution. For URW dynamics, we achieve an $O(\varepsilon^2)$ error with an iteration complexity of  $\tilde O(d^2/\varepsilon^2)$, significantly improving the $d$ and $m$ dependencies found in prior work. Furthermore, we derive TV distance bounds relative to the true target distribution, addressing a key gap in existing literature. However,
several open questions remain. It would be valuable to complement our analysis with a statistical study of DFM for specific neural network architectures as in  \citep{su2025theoreticalanalysisdiscreteflow, wan2025error}. 

\section*{Acknowledgments and Disclosure of Funding}
The work of LTNP has been supported by the \'Ecole Doctorale de Math\'ematiques Hadamard (EDMH). AD is supported by the France 2030 program (ANR-25-PEIA-0001, THEOREM project), Hi! Paris, and the Agence Nationale de la Recherche (Grant 11-LABX-0047). AD also received funding from the Fondation de l'\'Ecole polytechnique (Servir la science campaign) and the European Union (ERC-2022-SYG-OCEAN-101071601). Views and opinions expressed are those of the author(s) only and do not necessarily reflect those of the European Union or the European Research Council Executive Agency; neither the European Union nor the granting authority can be held responsible for them.

\bibliographystyle{alpha}
	\bibliography{ref_neurips}






\appendix

\section{Continuous Time Markov Chains and Jump Markov Bridge}\label{app:ctmc}

\paragraph{Continuous Time Markov Chains.}
Here we present briefly Continuous-Time Markov Chains (CTMCs) as they
are at the basis of the DFMs that we consider in this paper.
A CTMC on $\Z^d_m$ is a
time-indexed right-continuous stochastic process $(X_t)_{t \geq 0}$ on a probability space $(\Omega,\mcf,\PP)$ that Markov,
\ie,
for any $0 \leq s < t$, almost surely, $\P(X_{t_n} = x_{t_n} | \mcf_s)= \P(X_{t_n} = x_{t_n} | X_{s})$,
where $(\mcf_t)_{t\geq 0}$ is the natural filtration associated with $(X_t)_{t\geq0}$.

To design CTMCs, one central object is a rate matrix, \ie, function $(x,t,\msb)\in \Z^d_m \times \R_+ \times 2^{\Z^d_m} \mapsto q_t(x,\msb)$; also referred to $Q$-function or generator. In particular, we consider the following assumption on $(q_t)_{t\geq 0}$.
\begin{assumption}
	\label{ass:ctcm}
	$(q_t)_{t\geq 0}$ is a stable conservative rate matrix, \ie, it satisfies the following properties:
	\begin{itemize}
		\item for all $(x,t) \in \Z^d_m \times \rset_+$, the function $q_t(x,\cdot)$ is a signed (discrete) measure on $\Z^d_m$ such that $q_t(x,\Z^d_m) = 0$ and $0 \leq q_t(x,\mathsf{B}\setminus \{x\}) <\infty $ for all $\mathsf{B} \subset \Z^d_m$;
		\item for all $\mathsf{B} \subset \Z^d_m$, the function $(x,t) \mapsto q_t(x,\mathsf{B})$ is measurable;
		\item for all $x \in \Z^d_m$, the singleton $\{x\}$ is $q$-bounded, \ie,  $\sup_{t\in \rset_+} (-q_t(x, \{x\}))  <\infty$.
	\end{itemize}
\end{assumption}

Under~\Cref{ass:ctcm}, the generator $(q_t)_{t\geq 0}$ allows to define a sub-Markov semigroup $\{p_{s,t}\,:\, 0\leq s <t \}$ on $\Z^d_m$ and  corresponding CTMCs,
\ie, for $0\leq s <t$, $p_{t|s}$ is a  transition sub-probability density and  the Chapman-Kolmogorov equation holds, \ie, for any
$0\leq s< u < t$ and $x_s,x_u,x_t\in\Z^d_m$,
\begin{equation*}
	\label{eq:chapman_kolmogorov}
	p_{t|s}(x_t|x_s) = (p_{u|s}p_{t|u})(x_t|x_s) \eqdef \sum_{x_u \in\Z^d_m} p_{u|s}(x_u|x_s) p_{t|u}(x_t|x_u) \eqsp.
\end{equation*}
In the case where for any $0<s<t$, $p_{t|s} = p_{t-s|0}$, then $\{p_{s,t}\,:\, 0\leq s <t \}$ is said to be a homogeneous semigroup. Otherwise it is said
to be inhomogeneous.

Here we suppose that there is no explosion which is equivalent to the fact that the family of semigroup $\{p_{s,t}\,:\, 0\leq s <t\}$ is in fact  Markov.
\begin{assumption}
	\label{ass:non_explo}
	For any $0\leq s <t$ and $x \in\Z^d_m$, $p_{s,t}(x,\Z^d_m) = 1$.
\end{assumption}
\begin{remark}
	\label{rem:non_explo}
	Note that \citep[Theorem 6]{feller1940integro} ensure  that~\Cref{ass:non_explo} holds if  there exists a measurable function $t\mapsto \phi_t$ such that $\sup_{x\in\Z^d_m} (-q_t(x, \{x\})) \leq \phi_t$ for any $t \geq 0$, and for a fixed $p >1$, $\int_{s}^t (\phi_u)^p \rmd u < +\infty$ for any $0\leq s < t$.
\end{remark}
Under \Cref{ass:ctcm} and \Cref{ass:non_explo}, for any initialization $X_0\sim \mu_0$, there exists a CTMC $(X_t)_{t \geq 0}$ starting from $X_0$ and associated with the family of transitions $\{p_{s,t}\,:\, 0 \leq s <t \}$:
\begin{equation*}
	p_{t|s}(x_t|x_s) =
	\begin{cases}
		\PP(X_{t} = x_{t} | X_s = x_s) & \text{ if $\PP(X_s=x_s) \neq 0$} \eqsp, \\
		\PP(X_{t} = x_{t}) & \text{ otherwise } \eqsp.
	\end{cases}
\end{equation*}
In addition, \citep[Theorem 4.3.]{feinberg2014solutions} shows that for all $(x,s) \in \Z^d_m \times \R_+$ and $\msb \subset \Z^d_m$ such that $\sup_{t \in \R_+, x\in \msb} (-q_t(x,\{x\}))< \infty$, the function $p_{s,t}(x, \msb)$ satisfies for almost $t>s$ the forward Kolmogorov equation:
\begin{equation}
	\label{eq:foward_kolm}
	\frac{\partial p_{s,t}}{\partial t} (x,\msb) = \int_{\msb} q_t(y,\{y\}) p_{s,t}(x,\rmd y)  + \int_{\Z^d_m} q_t(y,\msb \setminus \{y\}) p_{s,t}(x,\rmd y)  \eqsp,
\end{equation}
and the backward Kolmogorov equation:
\begin{equation}\label{eq:backward_kolm}
	\frac{\partial p_{s,t}}{\partial s} (x,\msb) = -q_s(x,\{x\})p_{t|s}(\msb|x)  - \int_{\msx\setminus \{x\}} q_s(x,y) p_{t|s}(\msb|\rmd y)  \eqsp.
\end{equation}
We note that while we restrict our presentation to a family of
inhomogeneous generators $(q_t)_{t\geq 0}$ indexed by $\rset_+$, the
same results and construction apply when  it is restricted to a finite interval $[0,1]$.

\paragraph{Jump Markov Bridge.}
Based on a probability measure $\Q^q$ on the set of càdlàg paths $\W = \mathbf{D}([0,1], \Z^d_m)$ , we define the bridge $\B^q$ associated with $\Q^q$ as the Markov kernel on $(\Z^d_m)^2 \times \W$, such that, for any measurable set $\msa \subset \W$, $\Q^q(\msa) = \sum_{x_0,x_1} \B^q(\msa|x_0,x_1)\Q^q_{0,1}(x_0,x_1) $ where $\Q^q_{0,1}$
  denotes the joint distribution of the endpoints $(X_0,X_1)$ under $\Q^q$ . The existence of such a kernel is guaranteed by the existence of regular conditional probabilities (see, e.g., \citep[Theorem 8.37]{klenke2008probability}). Formally, for any finite set of time indices $\msi = \{t_1, \ldots, t_n\}$, we denote by $\Q^q_{\msi}$
  the $\msi$-marginal distribution of $\Q^q$
 , i.e., the pushforward measure of $\Q^q$
  under the evaluation map $(x_t)_{t\in [0,1]} \mapsto (x_{t_1}, \ldots, x_{t_n})$. From a probabilistic perspective, $\B^q$  represents the conditional distribution of the process $(X_t)_{t\in [0,1]}$
  given its endpoints: for any bounded measurable function $f$ on $\W$, $\E [f((X_t)_{t\in [0,1]})|X_0,X_1] = \int_{\W} f((x_t)_{t\in [0,1]}) \B^q(\rmd (x_t)_{t\in [0,1]}|X_0,X_1)$.

\section{Technical Proofs}\label{appendix:technical_proof}

\subsection{Proof of \eqref{eq:transition_density_nnrw}}\label{sec:poisson_expression_nnrw}
We first derive the expression for the one-dimensional case and subsequently extend the result to $d$ dimensions, leveraging the mutual independence of the coordinate processes.

We denote by $q^{(1)}$ the one-dimensional generator for any coordinate process $(X^i_t)_{t\in [0,1]}$, where $i \in [d]$ and $0 \leq t \leq 1$, which admits the following formula for $a, b \in \Z_m$:
\begin{align*}
	q^{(1)}(a,b) = \begin{cases}
		1/2 \quad &\text{if $a= b \pm 1$} \eqsp, \\
		-1 \quad &\text{if $a=b$} \eqsp, \\
		\, \,\, 0 \quad &\text{otherwise} \eqsp.
	\end{cases}
\end{align*}
We verify the identity $X_{t+s}^i - X_s^i = S_t^i \pmod m$ for $0\leq s <t \leq 1$ by demonstrating that the kernel
\[
p_{t+s|s}^{(1)}(b|a) = \sum_{k \in \mathbb{Z}} \mathbb{P}(S_t^i = b-a+km) = \rme^{-t}\sum_{k \in \Z} I_{|b-a+km|}(t)
\]
is the transition density for each component $i \in [d]$. This is achieved by proving that $p_{t+s|s}^{(1)}$ satisfies the one-dimensional Kolmogorov forward equation \eqref{eq:foward_kolm}. Indeed, for any $a \neq b$, we have
\begin{align*}
	\partial_t p_{t+s|s}^{(1)}(b|a) = - \rme^{-t} \sum_{k\in \Z} I_{|b-a+km|}(t) + \rme^{-t}\sum_{k\in \Z} \partial_t I_{|b-a+km|}(t) \eqsp,
\end{align*}
where we moved the derivative and the generator inside the summation as the sum converges absolutely and uniformly for $t > 0$.
Substituting the identity \( \partial_t I_n(t) = (I_{n-1}(t)+I_{n+1}(t))/2\) for $n \in \N^*$ yields
\begin{align*}
	\partial_t p_{t+s|s}^{(1)}(b|a) &= - \rme^{-t} \sum_{k\in \Z} I_{|b-a+km|}(t) + \frac{1}{2}\rme^{-t}\sum_{k\in \Z}(I_{|b-a+km|+1} + I_{|b-a+km|-1}) \\
	&= \sum_{c\in \Z_m}\sum_{k\in \Z} \rme^{-t}I_{|c-a+km|}(t) q^{(1)}(c,b) = \sum_{c\in \Z_m} p_{t+s|s}^{(1)}(c|a)q^{(1)}(c,b) \eqsp,
\end{align*}
where we used $I_{|z|+1}+ I_{|z|-1} = I_{|z+1|}+I_{|z-1|}$ for any $z \in \Z^*$ in the second equality.
Turning to the case where $a=b$, we obtain
\begin{align*}
	\partial_t p_{t+s|s}^{(1)}(a|a) &= -\partial_t \l(\sum_{b \neq a} p_{t+s|s}^{(1)}(b|a) \r)= - \sum_{b \neq a} \sum_{c\in \Z_m}   p_{t+s|s}^{(1)}(c|a)q^{(1)}(c,b) \\
	&=  -\sum_{c\in \Z_m}   p_{t+s|s}^{(1)}(c|a) \sum_{b \neq a}q^{(1)}(c,b) = \sum_{c\in \Z_m}   p_{t+s|s}^{(1)}(c|a) q^{(1)}(c,a) \eqsp,
\end{align*}
where the last line follows from the fact that $\sum_{b \in \Z_m} q^{(1)}(c,b) =0$ for any $c \in \Z_m$.
 The $d$-dimensional extension follows by showing that the product transition density $p_{t+s|s}(y|x) = \prod_{i=1}^d p_{t+s|s}^{(1)}(y^i|x^i)$ satisfies the global Kolmogorov forward equation with generator $q$ defined in \eqref{def:generator_nnrw}. Indeed, for $x,y \in \Z^d_m$, we have
\begin{align*}
	\partial_t p_{t+s|s}(y|x) = \partial_t \l(\prod_{i=1}^d p_{t+s|s}^{(1)}(y^i|x^i) \r)= \sum_{\ell =1}^d \partial_t p_{t+s|s}^{(1)}(y^\ell|x^\ell) \prod_{i\neq \ell}  p_{t+s|s}^{(1)}(y^i|x^i) \eqsp.
\end{align*}
As shown before, the one-dimensional kernel $p_{s+t|s}^{(1)}$ satisfies the following for any $a,b \in \Z_m$:
\begin{align*}
	\partial_t p_{t+s|s}^{(1)}(b|a) = \sum_{c\in \Z_m} p_{t+s|s}^{(1)}(c|a)q^{(1)}(c,b) \eqsp,
\end{align*}
which in turn implies
\begin{align*}
	\partial_t p_{t+s|s}(y|x)  &= \sum_{\ell=1}^d \sum_{j\in \Z_m}   p_{t+s|s}^{(1)}(j|x^\ell) q^{(1)}(j,y^\ell) \prod_{i\neq \ell}  p_{t+s|s}^{(1)}(y^i|x^i) \\
	&= \sum_{\ell=1}^d \sum_{j=-1}^{1} p_{t+s|s}^{(1)}(y^\ell+j|x^\ell) q^{(1)}(y^\ell+j,y^\ell)   \prod_{i\neq \ell}  p_{t+s|s}^{(1)}(y^i|x^i) \\
	&= \sum_{\sigma\in \Mc}  p_{t+s|s}(\sigma(y)|x)q(\sigma(y),y) + p_{t+s|s}(y|x)q(y,y) \\
	&= \sum_{z\in \Z^d_m} p_{t+s|s}(z|x) q(z,y) \eqsp.
\end{align*}
Consequently, $p_{t+s|s}$ satisfies the Kolmogorov forward equation associated with the nearest-neighbor random walk generator $q$. Invoking the uniqueness of solutions to the Kolmogorov equation \citep[Theorem 2.8.3]{norris1998markov}, we conclude that $p_{t+s|s}$ is the true transition density of $(X_t)_{t\in [0,1]}$, thereby validating its representation via the Poisson process.

\subsection{Proof of \Cref{theo:markov_projection}: Markovian Projection}\label{proof_theo:markov_projection}
	 For fixed $x_0, x_1 \in \Z^d_m$ and $0 \leq t <t+ \Delta t \leq 1$, we have
	\begin{align*}
		 	\P(X_{t+\Delta t}=y\mid X_t=x,\ X_1=x_1)
		&= \frac{p_{t+\Delta t|t}(y|x) p_{1|t+\Delta t}(x_1|y)}{p_{1|t}(x_1|x)} \eqsp.
	\end{align*}
	Recall that by Kolmogorov equation,
	\(p_{t+\Delta t|t}(y|x)=\mathbf 1_{y=x}+q(x,y)\Delta t+o(\Delta t)\) with $q$ is the homogeneous generator of $(X_t)_{t\in [0,1]}$ and analogously for the bridge generator $q_t^{x_0 \to x_1}$ between $x_0$ and $x_1$. Plugging this into the preceding equation yields
	\begin{align*}
	q^{x_0 \to x_1}_t (x,y)\Delta t + o(\Delta t) =  q(x,y)  \frac{p_{1|t+\Delta t}(x_1|y)}{p_{1|t}(x_1|x)}\Delta t + o(\Delta t) \eqsp,
 	\end{align*}
	for $x \neq y$. Divide both hand sides by $\Delta t$ and let $\Delta t \to 0$, we obtain that
	\[
		q^{x_0 \to x_1}_t(x,y) = q(x,y)  \frac{p_{1|t}(x_1|y)}{p_{1|t}(x_1|x)} \eqsp.
	\]
	Note that because the expression for $q^{x_0 \to x_1}_t$ is independent of the starting point $x_0$, it can be written more simply as $q^{(x_1)}_t$. Using this notation, we define the transition rate for $x \neq y$ over the interval $t \in [0,1]$ as:
	\begin{align*}
		\mq_t(x,y)\ :=\ \mathbb E\l[\,q_t^{(\iX_1)}(x,y)\ \big|\iX_t=x\,\r] \eqsp,
	\end{align*}
	with the diagonal terms $\mq_t(x,x) = -\sum_{y\neq x} \mq_t(x,y) $ chosen to enforce the row-sum constraint.
	A direct calculation yields the explicit form of the generator $\mq_t(x,y)$ for $y \neq x$:
	\begin{align*}
		\mq_t(x,y) =q(x,y) \uM_t(x,y) \eqsp,
	\end{align*}
	where the discrete score function $\uM$ is defined as
	\begin{align*}
		\uM_t(x,y) \eqdef \frac{\sum_{x_0,x_1\in \Z^d_m} p_{1|t}(x_1|y)p_{t|0}(x|x_0) \tilde\pi(x_0,x_1)}{\pI_t(x)} \eqsp,
	\end{align*}
    where $\tilde{\pi}(x_0, x_1) = \pi(x_0,x_1) / p_{1|0}(x_1|x_0)$.
	We now verify that the projected generator $\mq_t$ induces a probability flow that preserves the marginals of the mixture bridge $(\iX_t)_{t\in [0,1]}$ by showing that $\pI_t$ satisfies the Kolmogorov forward equation associated with $\mq_t$.
	Recall that $(s,t,x,y) \mapsto p_{t|s}(y|s)$ satisfies for any $x,y \in \Z^d_m$ and $0 \leq s <t \leq 1$ both the Kolmogorov forward and backward equation \eqref{eq:foward_kolm}, \eqref{eq:backward_kolm}.
	Exploiting these equations and \eqref{def:bridge_density}, for $t \in (0,1)$, we get that
	\begin{align*}
		\partial_t \pI_t(x) &= \sum_{x_0, x_1 \in \Z^d_m} \tilde{\pi}(x_0, x_1) \l[\partial_t p_{t|0}(x|x_0)p_{1|t}(x_1|x) +   p_{t|0}(x|x_0)\partial_t p_{1|t}(x_1|x) \r] \\
		&= \sum_{x_0, x_1 \in \Z^d_m} \tilde{\pi}(x_0, x_1) \sum_{z\in \Z^d_m} \l[p_{t|0}(z|x_0) q(z,x) p_{1|t}(x_1|x) - q(x,z) p_{1|t}(x_1|z)p_{t|0}(x|x_0) \r] \\
		&= \sum_{x_0, x_1 \in \Z^d_m} \tilde{\pi}(x_0, x_1) \sum_{z\neq x} \l[p_{t|0}(z|x_0) q(z,x) p_{1|t}(x_1|x) - q(x,z) p_{1|t}(x_1|z)p_{t|0}(x|x_0) \r]\\
		&=\sum_{z\neq x} \pI_t(z) \mq_t(z,x) - \pI_t(x) \underbrace{\sum_{z\neq x} \mq_t(x,z)}_{-\mq_t(x,x)} = \sum_{z\in \Z^d_m} \pI_t(z) \mq_t(z,x) \eqsp.
	\end{align*}
	Thus, the process $(\mX_t)_{t\in [0,1]}$ induced by $\mq$ matches the marginals of the mixture bridge $(\iX_t)_{t\in [0,1]}$. Since $(\mX_t)_{t
    \in [0,1]}$
  is Markovian by construction, and given that the linearity of the Kolmogorov forward equation ensures the uniqueness of the generator $\mq$
  within the support of the marginals, the Markovian projection is unique in law. This concludes the proof of the results in \Cref{theo:markov_projection}.

\subsection{Proof of \Cref{prop:bound_score}: Stability of the Score}\label{proof_prop:bound_score}
	Recall that for any $x \neq y$ and $t \in [0,1-\eta]$,
\begin{align*}
	\uM_t (x, y) = \E \l[ u_t^{(\iX_1)}(x, y) \middle| \iX_t =x \r] \eqsp,
\end{align*}
where $u_t^{(x_1)}$ denotes the bridge score and admits the following formula
\begin{align*}
	u_t^{(x_1)}(x, y) = \frac{p_{1|t}(x_1|y)}{p_{1|t}(x_1|x)} \eqsp.
\end{align*}
Therefore, it suffices to bound $u_t^{(x_1)}(x, y)$ to get the uniform bound of the score $\uM_t(x, y)$. As $q(x,y) =0$ for $y \notin \{x, \sigma(x): \sigma \in \Mc \}$, it suffices to focus on the case $y = \sigma(x)$ with $\sigma \in \Mc$.
We now evaluate the NNRW and URW cases individually to establish their respective bounds.

\paragraph{Nearest-Neighbor Random Walk.}
We begin by recalling
\begin{align*}
	u_t^{(x_1)}(x, \sigma^\ell_{\pm}(x)) &= \frac{p_{1|t}(x_1| \sigma^\ell_{\pm}(x) )}{p_{1|t}(x_1|x)}
	\overset{\eqref{eq:transition_density_nnrw}}{=}
	\frac{p^{(1)}_{1|t}(x_1^\ell| x^\ell \pm 1)}{p^{(1)}_{1|t}(x^\ell_1|x^\ell)}
	= \frac{\sum_{k \in \Z} I_{|\Delta \pm 1 + km|} (1-t)}{\sum_{k \in \Z} I_{|\Delta  + km|} (1-t)}
	\eqsp,
\end{align*}
where $\Delta \eqdef x_1^\ell - x^\ell$. Let us investigate $\sum_{k \in \Z} I_{|a+km|} (t)$ for $a \in \Z$, $t \in [0,1-\eta]$ to derive the bound on the interpolated score $	u_t^{(x_1)}(x, \sigma^\ell_{\pm}(x))$.
Let
\[
\tilde a = \min_{k \in \Z} |a+ km | \leq \min_{k \in \Z} (|a| + |k|m)  \leq \min_{k \in \Z} (1 + |k|)m -1 =m -1  \eqsp.
\]
We write
\begin{align}\label{eq:upper_bound_I}
	\sum_{k \in \Z} I_{|a+km|} (t)  = I_{\tilde a}(t) \sum_{k \in \Z} \frac{I_{|a+km|} (t)}{I_{\tilde a}(t) } \eqsp.
\end{align}
For $|a+km| - \tilde a \geq 1$, we have
\begin{align*}
	\frac{I_{|a+km|} (t)}{I_{\tilde a}(t) }  &= \prod_{j=0}^{|a+km| - \tilde a -1} \frac{I_{\tilde a +j+1}(t)}{I_{\tilde a +j } (t)} = \frac{I_{\tilde a + 1} (t)}{I_{\tilde a}(t)} \prod_{j=1}^{|a+km| - \tilde a -1} \frac{I_{\tilde a +j+1}(t)}{I_{\tilde a +j } (t)} \eqsp.
\end{align*}
Note that $I_{b}(t)$ is decreasing in $b \in \Z^+$ for fixed $t \in [0,1-\eta]$, thus $I_{\tilde a +1}(t) \leq I_{\tilde a}(t)$ and
\begin{align*}
	\frac{I_{|a+km|} (t)}{I_{\tilde a}(t) } \leq \prod_{j=1}^{|a+km| - \tilde a -1} \frac{I_{\tilde a +j+1}(t)}{I_{\tilde a +j } (t)} =  \prod_{j=1}^{|a+km| - \tilde a -1} R^+_{\tilde a +j}(t) \eqsp,
\end{align*}
where
\begin{align*}
	R^+_b(t) \eqdef \frac{I_{b+1}(t)}{I_b(t)} \quad \text{for $b \in \Z^+$} \eqsp.
\end{align*}
By Amos inequality \citep[Equation (9)]{amos1974computation}, we have
\begin{align*}
	R^+_{\tilde a +j}(t)  \leq \frac{t}{\tilde a +j + \sqrt{t^2+ (\tilde a+j)^2 }} \leq \frac{t}{2(\tilde a+j)} \overset{j \geq 1}{\leq} \frac{t}{2(\tilde a +1)} \eqsp.
\end{align*}
Hence for $|a+km| - \tilde a \geq 1$,
\begin{align*}
	\frac{I_{|a+km|} (t)}{I_{\tilde a}(t) } \leq \prod_{j=1}^{|a+km| - \tilde a -1} \frac{t}{2(\tilde a +1)} = \l(\frac{t}{2(\tilde a +1)} \r)^{|a+km| - \tilde a -1} \eqsp.
\end{align*}
Plugging it into \eqref{eq:upper_bound_I} gives
\begin{align*}
	\sum_{k \in \Z} I_{|a+km|} (t)  \leq & I_{\tilde a}(t) \l[1+ \sum_{\substack{k \in \Z \\ |a+km|- \tilde a \geq 1}} \l(\frac{t}{2(\tilde a +1)} \r)^{|a+km| - \tilde a -1} \r]\\
	\leq & I_{\tilde a}(t) \l[1+
	2	\sum_{k=0}^{\infty} \l(\frac{t}{2(\tilde a +1)} \r)^{k} \r]
	= I_{\tilde a}(t) \l[1+ \frac{2}{1-\frac{t}{2(\tilde a +1)}} \r]\\
	\leq & I_{\tilde a}(t) \Big[1+ \underbrace{\frac{4(\tilde a+1)}{2(\tilde a+1) -1}}_{\leq 4} \Big] \leq 5I_{\tilde a}(t) \eqsp.
\end{align*}
This together with the lower bound $\sum_{k \in \Z} I_{|a+km|} (t) \geq I_{\tilde a}(t)$ imply
\begin{align*}
	I_{\tilde a}(t) \leq 	\sum_{k \in \Z} I_{|a+km|} (t)  \lesssim I_{\tilde a}(t) \eqsp.
\end{align*}
As a result,
\begin{align*}
	\underbrace{\frac{ I_{\min_k |\Delta +km \pm 1|}(1-t)}{ I_{\min_k |\Delta +km|}(1-t)}}_{R}
	\lesssim \underbrace{\frac{\sum_{k \in \Z} I_{|\Delta \pm 1 + km|} (1-t)}{\sum_{k \in \Z} I_{|\Delta  + km|} (1-t)}}_{=u_t^{(x_1)} (x, \sigma^\ell_{\pm}(x))}
	\lesssim \frac{ I_{\min_k |\Delta +km \pm 1|}(1-t)}{ I_{\min_k |\Delta +km|}(1-t)}
\end{align*}
We evaluate the ratio $R$ by leveraging the decreasing property of $I_b(1-t)$ in $b \in \Z^+$ for fixed $t \in [0,1]$ as follows
\begin{align*}
	R &\geq \frac{I_{\min_k |\Delta+km| +1}(1-t)}{I_{\min_k |\Delta+km|}(1-t)} = R^+_{\min_k |\Delta +km|}(1-t) \\
	&\geq \frac{1-t}{\min_k|\Delta+km|+1+ \sqrt{(1-t)^2+ (\min_k|\Delta+km|+1)^2}} \\
	&\geq \frac{1-t}{m+\sqrt{(1-t)^2+m^2}} \geq \frac{1-t}{2m+1} \gtrsim \frac{1-t}{m } \eqsp,
\end{align*}
where we used Amos inequality \citep[Equation (9)]{amos1974computation} in the second line. Analogously, we can obtain the upper bound of $R$:
\begin{align*}
	R \leq \frac{ I_{\min_k |\Delta +km \pm 1|}(1-t)}{ I_{\min_k |\Delta +km \pm 1|+1}(1-t)} = \frac{1}{R^+_{\min_k |\Delta +km \pm 1|}(1-t)} \lesssim \frac{m}{1-t} \eqsp.
\end{align*}
Altogether yields the final bound on $u_t^{(x_1)} (x, \sigma^\ell_{\pm}(x))$:
\begin{align*}
	\frac{1-t}{m}  \lesssim u_t^{(x_1)} (x, \sigma^\ell_{\pm}(x)) \lesssim \frac{m }{1-t}  \eqsp.
\end{align*}
By taking the conditional expectation, we deduce the same uniform bound on $\uM_t(x,\sigma^\ell_{\pm}(x))$:
\begin{align*}
	\frac{1-t}{m} \lesssim \uM_t(x,\sigma^\ell_{\pm}(x)) \lesssim \frac{m }{1-t}  \eqsp.
\end{align*}
Moreover, the bound of the bridge score also implies
\begin{align*}
    \sum_{\sigma \in \Mc} u_t^{(x_1)} (x, \sigma(x)) = \sum_{\ell=1}^d  u_t^{(x_1)} (x, \sigma^\ell_{+}(x))+ \sum_{\ell=1}^d  u_t^{(x_1)} (x, \sigma^\ell_{-}(x)) \lesssim \frac{dm}{1-t} \eqsp.
\end{align*}
Taking the conditional expectation leads to an identical bound for $\sum_{\sigma \in \Mc}\uM_t(x,\sigma(x))$.

\paragraph{Uniform Random Walk.}
We begin by recalling
\begin{align}\label{eq:bridge_score_urw}
	u_t^{(x_1)}(x, y) &= \frac{p_{1|t}(x_1|y)}{p_{1|t}(x_1|x)}
	\overset{\eqref{eq:transition_density_urw}}{=}\frac{\alpha_{1-t}^{d_{\mrh}(y,x_1)}}{\alpha_{1-t}^{d_{\mrh}(x,x_1)}} = \alpha_{1-t}^{d_{\mrh}(y,x_1) - d_{\mrh}(x,x_1)} \eqsp.
\end{align}
By triangle inequality, we get that
\begin{align*}
	d_{\mrh}(x,x_1) - d_{\mrh}(y,x) \leq d_{\mrh} (y,x_1) \leq d_{\mrh}(x,x_1) + d_{\mrh} (x,y) \eqsp,
\end{align*}
which implies the following estimate since $\alpha_t < 1$ for any $t \in [0,1-\eta]$:
\begin{align}\label{eq:bound_score_1}
{ \quad
		\alpha_{1-t}^{-d_{\mrh}(y,x)} \geq  u_t^{(x_1)}(x, y) \geq \alpha_{1-t}^{d_{\mrh}(y,x)} \eqsp. \quad
	}
\end{align}
For $y = \sigma^\ell_n(x)$ with $\ell \in [d], n \in [m-1]$, this estimate reduces to
\begin{align*}
	\alpha_{1-t}^{-1} \geq u_t^{(x_1)}(x, \sigma^\ell_n(x)) \geq \alpha_{1-t} \eqsp,
\end{align*}
which yields the same bound on $\uM_t(x,\sigma^\ell_n(x))$.
Note that $\alpha_{1-t}^{-1}$ admits the following upper bound
\begin{align}\label{eq:bound_alpha_urw}
	\alpha_{1-t}^{-1} = \frac{1+(m-1)\rme^{-(1-t)}}{1-\rme^{-(1-t)}} = \frac{\rme^{1-t} + m-1}{\rme^{1-t}-1} \leq \frac{\rme +  m-1}{1-t} \lesssim \frac{m}{1-t}\eqsp,
\end{align}
where we used the inequality $\rme^x \geq x+1$ for any $x \in \R$ together with the fact $m \geq 2$. Thus
\begin{align*}
	{ \quad
		\frac{1-t}{m} \lesssim \uM_t(x, \sigma^\ell_n(x)) \lesssim \frac{m}{1-t} \eqsp.
		\quad }
\end{align*}
Furthermore, \eqref{eq:bridge_score_urw} implies
\begin{align*}
    \sum_{\sigma \in \Mc} u_t^{(x_1)}(x, \sigma(x)) &= \sum_{\ell=1}^d \sum_{n=1}^{m-1} \alpha_{1-t}^{d_{\mrh}(x_1, \sigma^\ell_n(x)) - d_{\mrh}(x_1,x)} \\
    &=  \sum_{\ell=1}^d \sum_{n=1}^{m-1}  \Big(\underbrace{\alpha_{1-t}}_{\leq 1} \1_{x^\ell_1=x^\ell}  + \alpha_{1-t}^{-1} \1_{x_1^\ell = (\sigma^\ell_n(x))^\ell} + \1_{x_1^\ell \notin \{x^\ell, (\sigma^\ell_n(x))^\ell \}}\Big) \\
    \leq & \sum_{\ell=1}^d \l( \alpha_{1-t}^{-1} + \sum_{n=1}^{m-1} \1_{x_1^\ell \neq (\sigma^\ell_n(x))^\ell} \r) \\
    &\overset{\eqref{eq:bound_alpha_urw}}{\lesssim} \sum_{\ell=1}^d \l( \frac{m}{1-t} + m-1 \r) \lesssim \frac{dm}{1-t} \eqsp.
\end{align*}
Taking the conditional expectation leads to an identical bound for $\sum_{\sigma \in \Mc}\uM_t(x,\sigma(x))$. The proof of \Cref{prop:bound_score} is complete.

\subsection{Proof of \Cref{prop:hjb}}\label{proof_prop:hjb}
Fix $\sigma \in \Mc$ and recall the following formula of the discrete score $\uM$  for $x \in \Z^d_m$, $t \in [0,1-\eta]$:
\[
\uM_t(x,\sigma(x)) = \frac{\sum_{x_0,x_1 \in \Z^d_m} p_{t|0}(x|x_0) p_{1|t}(x_1|y) \tilde \pi (x_0,x_1)}{\pI_t(x)} \eqsp,.
\]
 By quotient rule, we get that
\begin{align*}
	 \partial_t \uM_t(x, \sigma(x))  &= \sum_{x_0,x_1 \in \Z^d_m} \frac{\tilde \pi(x_0, x_1)}{\pI_t(x)}
	  \big[p_{1|t}(x_1|\sigma(x)) \partial_t p_{t|0}(x|x_0) +   p_{t|0}(x|x_0) \partial_t p_{1|t}(x_1| \sigma(x)) \big]\\
	  & \qquad - \sum_{x_0,x_1 \in \Z^d_m} \frac{\tilde \pi(x_0, x_1)}{(\pI_t(x))^2} p_{t|0}(x|x_0) p_{1|t}(x_1|\sigma(x))  \partial_t \pI_t(x) \eqsp.
\end{align*}
Employing the Kolmogorov forward and backward equations \eqref{eq:foward_kolm}, \eqref{eq:backward_kolm} satisfied by $p_{t|0}$, $p_{1|t}$ and $\pI_t$, we arrive at
\begin{align*}
	&\qquad \partial_t \uM_t(x, \sigma(x))\\
	  &= \lambda(m) \sum_{\sigma' \in \Mc} \sum_{x_0,x_1 \in \Z^d_m}\frac{\tilde \pi(x_0, x_1)}{\pI_t(x)} \Bigg[p_{1|t}(x_1|\sigma(x)) p_{t|0}(\sigma'(x)|x_0) - p_{1|t}(x_1|\sigma(x)) p_{t|0}(x|x_0) \\
	  &\qquad \qquad \qquad \qquad \qquad \qquad \qquad -p_{t|0}(x|x_0) p_{1|t}(x_1 | \sigma'(\sigma(x))) + p_{1|t}(x_1|\sigma(x)) p_{t|0}(x|x_0)\\
	  &\qquad \qquad \qquad \qquad \qquad  + p_{t|0}(x|x_0) p_{1|t}(x_1|\sigma(x)) \l( \uM_t(x, \sigma'(x))-\frac{\pI_t(\sigma'(x))}{\pI_t(x)} \uM_t(\sigma'(x),x) \r) \Bigg] \\
	  &= \lambda(m) \sum_{\sigma' \in \Mc} \Bigg\{ \uM_t(x, \sigma'(x)) \uM_t(x, \sigma(x)) - \uM_t(x, \sigma'(\sigma(x)))  \\
	  &\qquad \qquad \qquad \qquad + \frac{\pI_t(\sigma'(x))}{\pI_t(x)} \l[\uM_t(\sigma'(x), \sigma(x)) - \uM_t(\sigma'(x),x) \uM_t(x, \sigma(x)) \r]
	  \Bigg\} \eqsp,
\end{align*}
where $\lambda(m)$ represents the fixed jump rate, taking the value $1/2$ in the case of NNRW and $1/m$ for URW. Denoting the operator $\mcam$ as in \Cref{prop:hjb} concludes the proof.

\subsection{Proof of \Cref{prop:evolution_score}: Score and Entropy Evolution}\label{proof_prop:evolution_score}
\paragraph{Score Evolution.}
Fixing $\sigma \in \Mc$, we apply Itô's formula to the process $f^\sigma(t, \mX_t)$, where $f^\sigma(t,x) \eqdef \uM_t(x, \sigma(x))$. Observing that $\mX_t = \mX_{t-}$ for Lebesgue-almost every $t \in [0,1-\eta]$, we find that
\begin{align*}
	& f^\sigma(t, \mX_t) - f^\sigma (s, \mX_s)- \int_s^t \l[ \partial_r f^\sigma(r, \mX_r)  +  (\mq f^\sigma)(r, \mX_r) \r] \rmd r
\end{align*}
is a local martingale on $0 \leq s < t \leq 1 -\eta$. In view of the finite state space and the fact that $\uM_t$ is uniformly bounded for all $t \in [0,1-\eta]$ (as shown in \Cref{prop:bound_score}), we conclude that this local martingale is integrable and thus a true martingale.
Denoting by $b_r^{\sigma}(\mX_r)$ the integrand above, we have
\begin{align*}
	b^{\sigma}_r(x) &= \partial_r \uM_r(x, \sigma(x))  +  \lambda(m)\sum_{\sigma' \in \Mc} \l[ \uM_r(\sigma' (x), \sigma(\sigma' (x))) - \uM_r(x, \sigma(x))  \r]
	\uM_r (x, \sigma' (x))  \\
	&\overset{\eqref{eq:hjb_u}}{=} \lambda(m)
	\sum_{\sigma' \in \Mc} \Bigg\{ \uM_r(x,\sigma(x)) \uM_r(x, \sigma' (x)) -\uM_r(x,\sigma' (\sigma(x))) \\
	&\qquad \qquad \qquad \qquad + \frac{\pI_r(\sigma' (x))}{\pI_r(x)} \l[ \uM_r(\sigma' (x), \sigma(x)) - \uM_r(x,\sigma(x)) \uM_r(\sigma' (x),x) \r]\\
	&\qquad \qquad \qquad \qquad \qquad  + \l[ \uM_r(\sigma' (x), \sigma(\sigma' (x))) - \uM_r(x, \sigma(x))  \r]
	\uM_r (x, \sigma' (x))  \Bigg\} \\
	&=  \lambda(m)\sum_{\sigma' \in \Mc} \Bigg\{  \uM_r (x, \sigma' (x))\uM_r(\sigma' (x), \sigma(\sigma' (x))) -\uM_r(x,\sigma' (\sigma(x))) \\
	&\qquad \qquad \qquad + \frac{\pI_r(\sigma' (x))}{\pI_r(x)} \l[ \uM_r(\sigma' (x), \sigma(x)) - \uM_r(x,\sigma(x)) \uM_r(\sigma' (x),x) \r] \Bigg\} \eqsp.
\end{align*}
Consequently, we obtain the desired conclusion: for $0 \leq s < t \leq 1 - \eta$,
\begin{equation*}
		\begin{aligned}
		 \E \l[\uM_t(\sigma) \middle| \Fc_s \r] - \uM_s(\sigma) =  \lambda(m)  \int_s^t \E \l[\sum_{\sigma' \in \Mc} \mcbm_r(\mX_r, \uM; \sigma, \sigma') \middle| \Fc_s\r] \rmd r \eqsp,
	\end{aligned}
\end{equation*}
where $(\Fc_t)_{t\in [0,1-\eta]}$ denotes the filtration of $(\mX_t)_{t\in [0,1-\eta]}$ and the operator $\mcbm$ defined as in \Cref{prop:evolution_score}.

\paragraph{Entropy Score Evolution.}
Next, we investigate the evolution of $\sum_{\sigma \in \Mc} \varphi(\uM_t(\sigma))$-a central component in the convergence derivation, where $\varphi(a) \eqdef a\log(a) -a+1$. Analogously as before, by It\^o's formula, the following process
\begin{align*}
	& \varphi(\uM_t(\sigma)) - \varphi(\uM_s(\sigma)) - \int_s^t \l[ \partial_r \varphi(\uM_r(\sigma)) + (\mq_r \varphi(\uM_r))(\mX_r, \sigma(\mX_r))   \r] \rmd r
\end{align*}
is a local martingale for $0 \leq s <t \leq 1-\eta$.
The finiteness of the state space, coupled with the uniform bound on $\uM_t$ provided in \Cref{prop:bound_score}, ensures that the local martingale is integrable. Consequently, it becomes a true martingale throughout $[0,1-\eta]$.
Denoting by $c_r^\sigma(\mX_r)$ the above integrand, we have
\begin{align*}
	 &\qquad c_r^\sigma(x)\\
	  &= \partial_r \varphi(\uM_r(x, \sigma(x)))  +  \lambda(m)\sum_{\sigma' \in \Mc} \l[ \varphi(\uM_r(\sigma' (x), \sigma(\sigma' (x)))) - \varphi( \uM_r(x, \sigma(x)) )  \r]
	\uM_r (x, \sigma' (x))  \\
	&=  \log \uM_r(x, \sigma(x)) \partial_r \uM_r(x, \sigma(x)) + \lambda(m) \sum_{ \sigma' \in \Mc} \uM_r (x, \sigma' (x)) \\
	&\qquad \qquad \qquad \qquad  \l[ (\uM_r \log \uM_r - \uM_r)(\sigma' (x), \sigma(\sigma' (x))) - (\uM_r \log \uM_r - \uM_r)(x, \sigma(x)) \r] \eqsp.
\end{align*}
Substituting the equation \eqref{eq:hjb_u} into the above yields
\begin{align*}
	&\qquad c_r^\sigma(x) \\
	&= \lambda(m) \log \uM_r(x, \sigma(x))\sum_{\sigma' \in \Mc} \Bigg\{ \uM_r(x,\sigma(x)) \uM_r(x, \sigma' (x)) -\uM_r(x,\sigma' (\sigma(x)))\\
	&\qquad \qquad \qquad \qquad \qquad \qquad \quad + \frac{\pI_r(\sigma' (x))}{\pI_r(x)} \l[ \uM_r(\sigma' (x), \sigma(x)) - \uM_r(x,\sigma(x)) \uM_r(\sigma' (x),x) \r]\Bigg\}  \\
	&+ \lambda(m) \sum_{\sigma, \sigma'} {\color{black}\uM_r (x, \sigma' (x)) } \l[ (\uM_r \log \uM_r - \uM_r)(\sigma' (x), \sigma(\sigma' (x))) - ({\color{black}\uM_r \log \uM_r} - \uM_r){\color{black}(x, \sigma(x))} \r]  \\
	&= \lambda(m) \sum_{\sigma' \in \Mc} \Bigg\{ \frac{\pI_r(\sigma' (x))}{\pI_r(x)} \log \uM_r(x, \sigma(x)) \l[ \uM_r(\sigma' (x), \sigma(x)) - \uM_r(x,\sigma(x)) \uM_r(\sigma' (x),x) \r] \\
	& \qquad \qquad \qquad \qquad + \uM_r(x,\sigma' (x)) \l[
	(\uM_r \log \uM_r - \uM_r)(\sigma' (x), \sigma(\sigma' (x))) + \uM_r(x,\sigma(x))
	\r] \\
	&\qquad \qquad \qquad \qquad \qquad \qquad - \uM_r(x,\sigma' (\sigma(x))) \log \uM_r(x, \sigma(x)) \Bigg\} \eqsp.
\end{align*}
Taking expectation and noting that $\mX_t \overset{\text{dist}}{=} \iX_t$, we arrive at
\begin{align*}
	&\qquad \E \l[c_r^\sigma(\mX_r) \r] \\
	&=  \lambda(m)\sum_{\sigma' \in \Mc} \sum_{x \in \Z^d_m} \Bigg\{ {\pI_r(\sigma' (x))} \log \uM_r(x, \sigma(x)) \l[ \uM_r(\sigma' (x), \sigma(x)) - \uM_r(x,\sigma(x)) \uM_r(\sigma' (x),x) \r] \\
	&\qquad \qquad \qquad \qquad+ \pI_r(x) \uM_r(x,\sigma' (x)) \l[
	(\uM_r \log \uM_r - \uM_r)(\sigma' (x), \sigma(\sigma' (x))) + \uM_r(x,\sigma(x))
	\r] \\
	&\hspace{4cm}- \pI_r(x)\uM_r(x,\sigma' (\sigma(x))) \log \uM_r(x, \sigma(x)) \Bigg\}  \eqsp.
\end{align*}
A change of variables in the first summation yields
\begin{align*}
	 & \E \l[c_r^\sigma(\mX_r) \r]
	=  \lambda(m)\sum_{\sigma' \in \Mc} \sum_{x\in \Z^d_m} \pI_r(x)\\
	&\qquad \qquad \Bigg\{
	\uM_r(x, \sigma(\sigma' (x))) \log \uM_r(\sigma' (x), \sigma(\sigma' (x))) - \uM_r(x, \sigma' (\sigma(x))) \log \uM_r(x, \sigma(x)) \\
	&\qquad \qquad \qquad  + \uM_r(x, \sigma' (x)) \l[ \uM_r(x, \sigma(x)) - \uM_r(\sigma' (x), \sigma(\sigma' (x))) \r] \Bigg\}  \\
    &= \lambda(m) \E \Bigg[\sum_{\sigma'\in \Mc} \Bigg\{
	\uM_r(\mX_r, \sigma(\sigma' (\mX_r))) \log \uM_r(\sigma' (\mX_r), \sigma(\sigma' (\mX_r))) \\
    &\qquad \qquad \qquad \qquad - \uM_r(\mX_r, \sigma' (\sigma(\mX_r))) \log \uM_r(\mX_r, \sigma(\mX_r)) \\
    &\qquad + \uM_r(\mX_r, \sigma' (\mX_r)) \l[ \uM_r(\mX_r, \sigma(\mX_r)) - \uM_r(\sigma' (\mX_r), \sigma(\sigma' (\mX_r))) \r] \Bigg\} \Bigg]
\end{align*}
As a result, for $0 \leq s <t \leq 1-\eta$, we obtain the desired equation:
\begin{equation*}
	\E\l[ \varphi(\uM_t(\sigma))\r] - \E \l[\varphi(\uM_s(\sigma)) \r] = \lambda(m) \int_s^t \sum_{\sigma' \in \Mc} \E \l[ \mccm_r(\mX_r, \uM;\sigma, \sigma') \r] \rmd r \eqsp,
\end{equation*}
where the operator $\mccm$ defined as in \Cref{prop:evolution_score}.

\subsection{Proof of main result: \Cref{theo:main_2}}\label{proof_theo:main_2}
We are now prepared to demonstrate our main convergence results.
Let $\{t_k\}_{k=0}^K$ be a partition of $[0, 1-\eta]$ with mesh size $h_{k+1} = t_{k+1} - t_k$. By combining the data processing inequality \citep[Lemma 1.6]{nutz2021introduction} with the Girsanov theorem for jump processes \citep[Theorem G.0.1]{conforti2025nonasymptoticconvergencediscretediffusion}, we have
\begin{align*}
		&\qquad \KL(\mu_{1-\eta}|\mathrm{Law}(\Xthetastar_{1-\eta}))\\
	\leq & \KL(\mathrm{Law}(\mX_{[0,1-\eta]}) | \mathrm{Law}(\Xthetastar_{[0,1-\eta]})) \notag \\
	&=  \sum_{k=0}^{K-1} \int_{t_k}^{t_{k+1}}  \E \Bigg[ \sum_{\sigma \in \Mc}  \l( {\mq_t}\log \frac{\mq_t}{\mqthetastar_{t}} - {\mq_t}+\mqthetastar_{t} \r)  (\mX_t,\sigma (\mX_t))  \Bigg] \rmd t \eqsp.
\end{align*}
Plugging the formula of estimated generator:  $\mqthetastar_t(x,\sigma(x)) = \mqthetastar_{t_k}(\mX_{t_k}, \sigma(\mX_{t_k}))$ for $t \in [t_k,t_{k+1})$ given $\mX_{t_k}$ and using the shorthand $\mq_t(\sigma) = \mq_t(\mX_t, \sigma(\mX_t))$ (and similarly for $\mq_{t_k}(\sigma), \mqthetastar_{t_k}(\sigma)$), we get that
\begin{align*}
    &\qquad \KL(\mu_{1-\eta}|\mathrm{Law}(\Xthetastar_{1-\eta}))\\
    \leq & \sum_{k=0}^{K-1} \int_{t_k}^{t_{k+1}}  \E \Bigg[ \sum_{\sigma \in \Mc}  \l( {\mq_t}\log \frac{\mq_t}{\mqthetastar_{t_k}} - {\mq_t}+\mqthetastar_{t_k} \r)  (\sigma)  \Bigg] \rmd t \\
    \leq & \sum_{k=0}^{K-1} \int_{t_k}^{t_{k+1}}  \E \Bigg[ \sum_{\sigma \in \Mc}  \Bigg( \mq_t\log \frac{\mq_t}{\mq_{t_k}} -\mq_t + \mq_{t_k} + \mqthetastar_{t_k} - \mq_{t_k} + {\mq_t}\log \frac{\mq_{t_k}}{\mqthetastar_{t_k}}  \Bigg)  (\sigma)  \Bigg] \rmd t \eqsp.
\end{align*}
We decompose this expression into the score approximation error and a residual discretization term:
\begin{align*}
    &\qquad \KL(\mu_{1-\eta}|\mathrm{Law}(\Xthetastar_{1-\eta}))\\
    \leq & \underbrace{\lambda(m)\sum_{k=0}^{K-1} \int_{t_k}^{t_{k+1}}  \E \Bigg[ \sum_{\sigma \in \Mc}  \Bigg( \varphi(\uM_t) -  \varphi(\uM_{t_k}) + \log \uM_{t_k}(\uM_{t_k}-\uM_t) \Bigg)  (\sigma)  \Bigg] \rmd t}_{E_{\text{disc}}}\\
    &+ \underbrace{\sum_{k=0}^{K-1} h_{k+1} \E \Bigg[ \sum_{\sigma \in \Mc} \l|\mqthetastar_{t_k}(\sigma) - \mq_{t_k}(\sigma) \r| \Bigg] + \sum_{k=0}^{K-1} \int_{t_k}^{t_{k+1}}  \E \Bigg[ \sum_{\sigma \in \Mc}
    \frac{\mq_t}{\mq_{t_k}} \mq_{t_k} \log \frac{\mq_{t_k}}{\mqthetastar_{t_k}} (\sigma)
    \Bigg] \rmd t}_{E_{\text{approx}}} \eqsp,
\end{align*}
where $\lambda(m) = q(x,\sigma(x))$ denotes the fixed jump rate taking $1/2$ for NNRW and $1/m$ for URW, and the nonnegative function $\varphi$ defined by $\varphi(a) = a\log(a)-a+1$ for $a \geq 0$.
To handle the approximation error $E_{\text{approx}}$, we first note that \Cref{ass:approx_error_2} is indeed equivalent to 
\begin{align}\label{eq:ass_error}
    \sum_{k=0}^{K-1} h_{k+1}  \E \l[\sum_{\sigma \in \Mc}\l\{ \mqtheta_{t_k} \varphi \l(\frac{\mq_{t_k}}{\mqtheta_{t_k}}\r) + \l( \mqtheta_{t_k} - \mq_{t_k} \r)^2 \r\} (\mX_{t_k}, \sigma(\mX_{t_k}) ) \r] \leq \tilde{\varepsilon}^2 \eqsp,
\end{align}
 since $\mX_t \overset{\text{dist}}{=} \iX_t$ for all $t \in [0,1]$. Now we address the second term of $E_{\text{approx}}$ by using the uniform bound of the score established in \Cref{prop:bound_score}, together with the note
\begin{align*}
    \l|\mq_{t_k} \log \frac{\mq_{t_k}}{\mqthetastar_{t_k}} (\sigma)\r|  \leq & \l|(\mq_{t_k} \log \frac{\mq_{t_k}}{\mqthetastar_{t_k}} - \mq_{t_k} + \mqthetastar_{t_k})(\sigma)\r| + \l|( \mq_{t_k} - \mqthetastar_{t_k})(\sigma)\r| \\
    &= \mqthetastar_{t_k} \varphi\l(\frac{\mq_{t_k}}{\mqthetastar_{t_k}} \r)(\sigma) + \l| \mq_{t_k}(\sigma) - \mqthetastar_{t_k}(\sigma)\r| \eqsp,
\end{align*}
where we used triangle inequality in the first estimate. Thereby
\begin{align*}
    E_{\text{approx}} &\lesssim \frac{m^2}{\eta^2} \sum_{k=0}^{K-1} h_{k+1} \E \l[ \sum_{\sigma\in \Mc}\l| \mq_{t_k}(\sigma) - \mqthetastar_{t_k}(\sigma)\r| \r] \\
    &\qquad \qquad +   \frac{m^2}{\eta^2} \underbrace{\sum_{k=0}^{K-1} h_{k+1} \E \l[ \sum_{\sigma\in \Mc} \mqthetastar_{t_k} \varphi\l(\frac{\mq_{t_k}}{\mqthetastar_{t_k}} \r)(\sigma) \r]}_{\leq \tilde{\varepsilon}^2 \text{ by \eqref{eq:ass_error}}} \eqsp.
\end{align*}
We first employ Cauchy-Schwarz inequality and second apply twice the Jensen inequality for concave functions to control the remaining term of $E_{\text{approx}}$:
\begin{align*}
    &\qquad \sum_{k=0}^{K-1} h_{k+1} \E \l[ \sum_{\sigma\in \Mc}\l| \mq_{t_k}(\sigma) - \mqthetastar_{t_k}(\sigma)\r| \r] \\
    \leq & \sum_{k=0}^{K-1} h_{k+1} \E \l[\sqrt{|\Mc|} \sqrt{\sum_{\sigma \in \Mc} (\mq_{t_k}(\sigma) - \mqthetastar_{t_k}(\sigma))^2}  \r] \\
    \leq & \sqrt{|\Mc|} \sum_{k=0}^{K-1} h_{k+1} \sqrt{\E \l[ \sum_{\sigma \in \Mc} (\mq_{t_k}(\sigma) - \mqthetastar_{t_k}(\sigma))^2\r]} \\
    \leq & \sqrt{|\Mc|}  \sqrt{\sum_{k=0}^{K-1} h_{k+1} \E \l[ \sum_{\sigma \in \Mc} (\mq_{t_k}(\sigma) - \mqthetastar_{t_k}(\sigma))^2\r]} \overset{\eqref{eq:ass_error}}{\leq}
    \sqrt{|\Mc|} \tilde{\varepsilon} \eqsp.
 \end{align*}
    Thus we obtain the following upper bound on $E_{\text{approx}}$:
\begin{align*}
    E_{\text{approx}} \lesssim m^2 \eta^{-2} \sqrt{|\Mc|}\tilde{\varepsilon} + m^2 \eta^{-2} \tilde{\varepsilon}^2 = m^2 \eta^{-2} (\sqrt{|\Mc|}\tilde{\varepsilon}+\tilde{\varepsilon}^2) \eqsp.
\end{align*}
It remains to control the time-discretization error $E_{\text{disc}}$ to finish the proof. We write
 \begin{align*}
 	E_{\text{disc}} &=  \lambda(m)\underbrace{\sum_{k=0}^{K-1} \int_{t_k}^{t_{k+1}}  \E \Bigg[ \sum_{\sigma \in \Mc}  \log \uM_{t_k} (\uM_{t_k} - \uM_t) (\sigma)\Bigg]  \rmd t}_{E_1} \\
 	&\qquad \qquad + \lambda(m)\underbrace{\sum_{k=0}^{K-1} \int_{t_k}^{t_{k+1}}    \E \Bigg[ \sum_{\sigma \in \Mc} \l\{ \varphi ( \uM_t(\sigma) ) - \varphi (\uM_{t_{k}}(\sigma)) \r\} \Bigg]  \rmd t}_{E_2}  \eqsp,
 \end{align*}

To analyze the term $E_2$, we use the evolution of the entropy-like score established in \Cref{prop:evolution_score}:
 \begin{align*}
 	& E_2 = \sum_{k=0}^{K-1} \int_{t_k}^{t_{k+1}}  \E \Bigg[ \sum_{\sigma \in \Mc} \l\{ \varphi ( \uM_t(\sigma) ) - \varphi (\uM_{t_{k}}(\sigma)) \r\} \Bigg] \rmd t \\
 	&\quad = \lambda(m)\sum_{k=0}^{K-1} \int_{t_k}^{t_{k+1}} \int_{t_k}^{t} \E \Bigg[ \sum_{\sigma, \sigma'} \Bigg\{\uM_r(\mX_r, \sigma(\sigma' (\mX_r))) \log \frac{\uM_r(\sigma' (\mX_r),\sigma(\sigma' (\mX_r)))}{\uM_r (\mX_r, \sigma(\mX_r))} \\
 	&\qquad \qquad \qquad+ \uM_r (\mX_r, \sigma' (\mX_r)) \bigg[ \uM_r(\mX_r, \sigma(\mX_r)) - \uM_r(\sigma' (\mX_r), \sigma(\sigma' (\mX_r)))  \bigg] \Bigg\} \Bigg] \rmd r \rmd t \\
 	&\quad \leq  \lambda(m) \underbrace{\sum_{k=0}^{K-1} \int_{t_k}^{t_{k+1}}  \int_{t_k}^{t_{k+1}} \E \Bigg[ \sum_{\sigma, \sigma'} \uM_r (\mX_r, \sigma' (\mX_r)) \uM_r(\mX_r, \sigma(\mX_r))   \Bigg] \rmd r \rmd t}_{E_{2.1}} \\
 	& +\lambda(m) \underbrace{\sum_{k=0}^{K-1} \int_{t_k}^{t_{k+1}}   \int_{t_k}^{t_{k+1}} \E \Bigg[ \sum_{\sigma, \sigma'}\uM_r(\mX_r, \sigma(\sigma' (\mX_r))) \l|\log \frac{\uM_r(\sigma' (\mX_r),\sigma(\sigma' (\mX_r)))}{\uM_r (\mX_r, \sigma(\mX_r))} \r| \Bigg] \rmd r \rmd t}_{E_{2.2}}\eqsp.
 \end{align*}
We address $E_{2.1}$ by the upper bound shown in \eqref{eq:bound_sum_score}:
\begin{align*}
    E_{2.1} &= \sum_{k=0}^{K-1} \int_{t_k}^{t_{k+1}}  \int_{t_k}^{t_{k+1}} \E \Bigg[ \sum_{ \sigma' \in \Mc} \uM_r (\mX_r, \sigma' (\mX_r)) \sum_{\sigma \in \Mc} \uM_r(\mX_r, \sigma(\mX_r))   \Bigg] \rmd r \rmd t \\
    &\lesssim \sum_{k=0}^{K-1} h_{k+1} \int_{t_k}^{t_{k+1}}\frac{d^2 m^2}{(1-r)(1-t_{k+1})} \rmd r \eqsp.
\end{align*}
Let us choose	\(
 t_k = 1 - (1+h)^{-k}
 \) with $h>0$, then $ h_{k+1} = h(1-t_{k+1}) $, thus
 \begin{align}\label{eq:e21}
     E_{2.1} &\lesssim d^2 m^2 h \int_0^{1-\eta} \frac{1}{1-r} \rmd r = d^2 m^2 h \log(\eta^{-1}) \eqsp.
 \end{align}
Next we invoke \Cref{prop:bound_score} to control the logarithmic term of $E_{2.2}$: for any $x, x' \in \Z^d_m $, $t, t_k \in  [0,1-\eta]$ and $\sigma\in \Mc$,
 \begin{align*}
 	\l|\log \frac{\uM_t(x,\sigma(x))}{\uM_{t_k} (x', \sigma(x'))} \r| \lesssim \log \l(\frac{m}{1-t} \r)  + \log \l(\frac{m}{1-t_k} \r)  \lesssim \log \l(\frac{m}{\eta} \r) \eqsp.
 \end{align*}
  In addition, \Cref{prop:bound_score} further implies for any $x \in \Z^d_m$ and $r \in [t_k,t_{k+1}]$,
\begin{align*}
    \sum_{\sigma, \sigma'} \uM_r(x,\sigma(\sigma'(x))) &= \sum_{\sigma, \sigma'} \E \l[u_r^{\iX_0 \to \iX_1} (x,\sigma(\sigma'(x))) \middle| \iX_r = x \r] \\
 	&= \E \l[\sum_{\sigma'\in \Mc} u_r^{\iX_0 \to \iX_1} (x, \sigma'(x)) \sum_{\sigma\in \Mc} u_r^{\iX_0 \to \iX_1} (\sigma'(x),\sigma(\sigma'(x))) \middle| \iX_r = x \r]\eqsp \\
    &\overset{\eqref{eq:bound_sum_score}}{\lesssim} \E \l[\frac{d^2m^2}{(1-r)^2} \middle| \iX_r=x\r] \leq \frac{d^2 m^2}{(1-r)(1-t_{k+1})} \eqsp,
\end{align*}
 where we employed the transitivity of the bridge score in the second equality: \[u_t^{(x_1)} (x,y) u_t^{(x_1)} (y,z)  = \frac{p_{1|t}(x_1|y)}{p_{1|t}(x_1|x)}\cdot\frac{p_{1|t}(x_1|z)}{p_{1|t}(x_1|y)} = \frac{p_{1|t}(x_1|z)}{p_{1|t}(x_1|x)} =u_t^{(x_1)} (x,z) \eqsp,\]
 for any $x,y,z \in \Z^d_m$ and $t\in [0,1-\eta]$. Therefore
 \begin{align}\label{eq:e22}
     E_{2.2} &\lesssim \log(m\eta^{-1})\sum_{k=0}^{K-1} h_{k+1} \int_{t_k}^{t_{k+1}}\frac{d^2 m^2}{(1-r)(1-t_{k+1})} \rmd r \notag\\
     &= d^2 m^2 h\log(m\eta^{-1})\int_0^{1-\eta}\frac{1}{1-r} \rmd r = d^2 m^2 h \log(\eta^{-1})\log(m\eta^{-1}) \eqsp.
 \end{align}
 Combining \eqref{eq:e21} and \eqref{eq:e22} yields
 \begin{align}\label{eq:e2}
     E_2 \lesssim \lambda(m) d^2 m^2 h \log(\eta^{-1})\log(m\eta^{-1}) \eqsp.
 \end{align}
To analyze $E_1$, we employ the evolution of the score showed in \Cref{prop:evolution_score} to get
 \begin{align*}
 	E_1 &= \lambda(m)\sum_{k=0}^{K-1} \int_{t_k}^{t_{k+1}} \int_{t_k}^t   \E \Bigg[  \sum_{\sigma, \sigma'} \log {\uM_{t_k}(\sigma)} \\
 	&\quad  \Big\{  \uM_r(\mX_r,\sigma' (\sigma(\mX_r))) - \uM_r(\mX_r,\sigma' (\mX_r)) \uM_r(\sigma' (\mX_r),\sigma( \sigma' (\mX_r))) \\
 	&\quad+ \frac{\pI_r(\sigma' (\mX_r))}{\pI_r(\mX_r)} \l[   \uM_r(\mX_r,\sigma(\mX_r)) \uM_r(\sigma' (\mX_r),\mX_r)  - \uM_r(\sigma' (\mX_r), \sigma(\mX_r)) \r] \Big\}  \Bigg] \rmd r  \rmd t \\
 	\leq & \lambda(m) \sum_{k=0}^{K-1} h_{k+1} \int_{t_k}^{t_{k+1}}   \E \Bigg[ \sum_{\sigma, \sigma'}  \l|  \log {\uM_{t_k}(\sigma)} \r| \\
 	&\qquad \Big\{ \l|  \uM_r(\mX_r,\sigma' (\sigma(\mX_r))) - \uM_r(\mX_r,\sigma' (\mX_r)) \uM_r(\sigma' (\mX_r),\sigma( \sigma' (\mX_r))) \r| \\
 	&\qquad+  \frac{\pI_r(\sigma' (\mX_r))}{\pI_r(\mX_r)} \l| \uM_r(\mX_r,\sigma(\mX_r)) \uM_r(\sigma' (\mX_r),\mX_r) - \uM_r(\sigma' (\mX_r), \sigma(\mX_r)) \r| \Big\} \Bigg] \rmd r \eqsp.
 \end{align*}
Then we employ the uniform bound of the score established in \Cref{prop:bound_score} to control $|\log \uM_{t_k}(\sigma)|$:
\begin{align*}
    E_1 &{\lesssim} \lambda(m)\log \l(m{\eta}^{-1} \r)\sum_{k=0}^{K-1} h_{k+1}\\
 	& \int_{t_k}^{t_{k+1}}    \E \l[\sum_{\sigma, \sigma'} \l| \uM_r(\mX_r,\sigma' (\sigma(\mX_r))) - \uM_r(\mX_r,\sigma' (\mX_r)) \uM_r(\sigma' (\mX_r),\sigma( \sigma' (\mX_r)))\r|\r] \\
 	&+  \underbrace{\E \l[\sum_{\sigma, \sigma'} \frac{\pI_r(\sigma' (\mX_r))}{\pI_r(\mX_r)} \l| \uM_r(\mX_r,\sigma(\mX_r)) \uM_r(\sigma' (\mX_r),\mX_r) - \uM_r(\sigma' (\mX_r), \sigma(\mX_r))  \r| \r] }_{E_{1.1}}\rmd r \eqsp.
\end{align*}

The addend $E_{1.1}$ can be controlled by changing variables as follows
 \begin{align*}
 	E_{1.1} &=\sum_{\sigma, \sigma'} \sum_{x \in \Z^d_m} \pI_r(\sigma' (x)) \l|  \uM_r (\sigma' (x), \sigma(x)) - \uM_r(x, \sigma(x)) \uM_r(\sigma' (x),x)  \r| \\
 	&= \sum_{\sigma, \sigma'}\sum_{x\in \Z^d_m} \pI_r(x) \l| \uM_r(x, \sigma(\sigma' (x))) - \uM_r(\sigma' (x), \sigma(\sigma' (x))) \uM_r(x, \sigma' (x))  \r| \\
 	&= \sum_{\sigma, \sigma'}\E \l[ \l| \uM_r(\mX_r,\sigma' (\mX_r)) \uM_r(\sigma' (\mX_r),\sigma( \sigma' (\mX_r))) - \uM_r(\mX_r,\sigma' (\sigma(\mX_r))) \r| \r] \eqsp,
 \end{align*}
 where we leveraged the commutative property of the set of jumps $\Mc$ ($\sigma \circ \sigma' = \sigma' \circ \sigma$) in the last equality. Substituting back into $E_1$ leads to:
 \begin{align*}
 	E_1 &\lesssim  \lambda(m)\log \l(m{\eta}^{-1} \r) \sum_{k=0}^{K-1} h_{k+1}\\
    & \int_{t_k}^{t_{k+1}} \E \l[\sum_{\sigma, \sigma'} \l|\uM_r(\mX_r,\sigma' (\mX_r)) \uM_r(\sigma' (\mX_r),\sigma( \sigma' (\mX_r)))  -\uM_r(\mX_r,\sigma' (\sigma(\mX_r))) \r|\r] \rmd r\\
    &\lesssim   \lambda(m)\log \l(m{\eta}^{-1} \r) \sum_{k=0}^{K-1} h_{k+1}\\
 	& \int_{t_k}^{t_{k+1}}    \E \l[ \sum_{\sigma, \sigma'} \l\{\uM_r(\mX_r,\sigma' (\sigma(\mX_r))) + \uM_r(\mX_r,\sigma' (\mX_r)) \uM_r(\sigma' (\mX_r),\sigma( \sigma' (\mX_r))) \r\} \r] \rmd r \eqsp.
 \end{align*}
 Arguing similarly as $E_2$, we obtain
\begin{align*}
    E_1 \lesssim \lambda(m)\log \l(m{\eta}^{-1} \r) \sum_{k=0}^{K-1} d^2 m^2 \frac{h_{k+1}}{1-t_{k+1}} \int_{t_k}^{t_{k+1}} \frac{1}{1-r} \rmd r \eqsp.
\end{align*}
Recall that $ h_{k+1} = h(1-t_{k+1}) $, we arrive at
\begin{align}\label{eq:e1}
    E_1 &\lesssim \lambda(m)\log \l(m{\eta}^{-1} \r) d^2 m^2 h \int_0^{1-\eta} \frac{1}{1-r} \rmd r \notag \\
    &= \lambda(m) d^2 m^2 h \log \l(m{\eta}^{-1} \r) \log \l({\eta}^{-1} \r) \eqsp.
\end{align}
Plugging the estimates for $E_1$ and $E_2$ in \eqref{eq:e2}, \eqref{eq:e1} into
the universal $E_{\text{disc}}$ yields
\begin{align*}
    E_{\text{disc}} \lesssim (\lambda(m))^2 d^2 m^2 h \log \l(m{\eta}^{-1} \r) \log \l({\eta}^{-1} \r)  \eqsp.
\end{align*}
Putting pieces together implies the final error bound
 \begin{align*}
 	\KL(\mu_{1-\eta}| \mathrm{Law}(\Xthetastar_{1-\eta})) &\lesssim   m^2 \eta^{-2} (\sqrt{|\Mc|}\tilde{\varepsilon}+\tilde{\varepsilon}^2) +  (\lambda(m))^2 d^2 m^2 h \log \l(m{\eta}^{-1} \r) \log \l({\eta}^{-1} \r) \eqsp,
 \end{align*}
and we complete the proof of \Cref{theo:main_2}.

\subsection{Proof of \Cref{cor:tv_distance_2}: Error in TV distance}\label{proof_cor:tv_distance_2}
We start by applying triangle inequality:
\begin{align}\label{eq:cor_main}
		\| \mathrm{Law}(\Xthetastar_{1-\eta}) - \mu_1 \|_{\TV} \leq & \| \mathrm{Law}(\Xthetastar_{1-\eta}) - \mu_{1-\eta} \|_{\TV}  + 		\| \mu_{1-\eta} - \mu_1 \|_{\TV} \eqsp.
\end{align}
The first addend can be bounded by leveraging Pinsker's inequality and \Cref{theo:main_2}:
\begin{align*}
		\| \mathrm{Law}(\Xthetastar_{1-\eta}) - \mu_{1-\eta} \|_{\TV}  \leq & \sqrt{2\KL (\mu_{1-\eta} | \mathrm{Law}(\Xthetastar_{1-\eta}))} \\
        &\lesssim  m\eta^{-1}\sqrt{\sqrt{|\Mc|} \tilde{\varepsilon} + \tilde{\varepsilon}^2 } + \lambda(m)d m\sqrt{h \log \l(m\eta^{-1}\r) \log(\eta^{-1})} \eqsp.
\end{align*}
where we used inequality $\sqrt{a+b} \leq \sqrt{a}+ \sqrt{b}$ for non-negative numbers $a,b$ in last estimate.
To conclude the proof, it remains to control the approximation error $\|\mu_{1-\eta} - \mu_1\|_{\TV}$ arising from the early-stopping procedure. This term represents the discrepancy between the regularized flow and the true target distribution; its analysis depends on the specific choice of base dynamics as follows.

\paragraph{Nearest-Neighbor Random Walk.}
We have
\begin{align*}
	\| \mu_{1-\eta} - \mu_1 \|_{\TV} \leq & \P \l( \iX_{1-\eta} \neq \iX_1 \r) = 1- \P(\iX_{1-\eta} = \iX_1) \\
	&= 1-\sum_{x_0, x_1}\pi(x_0,x_1)  \frac{p_{1-\eta|0}(x_1|x_0) p_{1|1-\eta} (x_1|x_1)}{p_{1|0}(x_1|x_0)}\eqsp.
\end{align*}
Thanks to the rigorous formula of transition density provided in \eqref{eq:transition_density_nnrw}, we get that
\begin{align}\label{eq:cor_1}
	p_{1|1-\eta}(x_1|x_1) &= \prod_{\ell=1}^d p^{(1)}_{1|1-\eta}(x_1^\ell|x_1^\ell) = \l(\rme^{-\eta} \sum_{k\in \Z} I_{|km|}(\eta) \r)^d \geq \rme^{-d\eta} I_0^d(\eta) \geq \rme^{-d\eta} \eqsp,
\end{align}
since
\[
I_0(\eta) = \sum_{n=0}^\infty \frac{1}{(n!)^2} \l(\frac{\eta}{2} \r)^{2n} \geq \frac{1}{(0!)^2} \l(\frac{\eta}{2}\r)^{0} = 1 \eqsp.
\]
The tricky part is to manage the ratio $y_\eta = p_{1-\eta|0}(x_1|x_0)/p_{1|0}(x_1|x_0)$. By Kolmogorov backward equation and homogeneity of the base process, we have
\begin{align*}
	y'_s &=  \frac{\partial_s p_{1|s}(x_1|x_0)}{p_{1|0}(x_1|x_0)} = -\sum_{x \in \Z^d_m} q(x_0,x) \frac{p_{1|s}(x_1|x)}{p_{1|0}(x_1|x_0)} \\
	&= d\frac{  p_{1|s}(x_1|x_0) }{p_{1|0}(x_1|x_0)}- \frac{1}{2} \sum_{\sigma \in \Mc} \frac{p_{1|s} (x_1|\sigma(x_0))}{p_{1|0}(x_1|x_0)} \\
	&= d y_s - \frac{1}{2}y_s \sum_{\sigma \in \Mc} \underbrace{\frac{p_{1|s}(x_1|\sigma(x_0))}{p_{1|s}(x_1|x_0)}}_{u_s^{x_0 \to x_1}(x_0, \sigma(x_0))} \eqsp.
\end{align*}
Thanks to the precise upper bound of the bridge score derived in the proof of \Cref{prop:bound_score}, we get that
\begin{align*}
	y'_s \geq dy_s -\frac{1}{2}y_s \sum_{\sigma\in \Mc} \frac{5(2m+1)}{1-s} = dy_s \l(1-\frac{5(2m+1)}{1-s} \r) \eqsp.
\end{align*}
Therefore
\begin{align*}
	\frac{y'_s}{y_s} \geq d\l[1-\frac{5(2m+1)}{1-s} \r] \eqsp.
\end{align*}
Integrating from $0$ to $\eta$ gives
\begin{align*}
	\log(y_\eta) -\log(y_0) \geq d\int_0^\eta \l(1-\frac{5(2m+1)}{1-s} \r) \rmd s = d [\eta +5(2m+1)\log(1-\eta) ] \eqsp.
\end{align*}
As $y_0 =1$, we deduce that
\begin{align}\label{eq:cor_2}
	y_\eta \geq \rme^{d\eta} (1-\eta)^{5d(2m+1)} \eqsp.
\end{align}
Plugging \eqref{eq:cor_1} and \eqref{eq:cor_2} into the total variation distance yields
\begin{align*}
	\|\mu_{1-\eta} - \mu_1\|_{\TV} \leq 1 - (1-\eta)^{5d(2m+1)}\sum_{x_0,x_1} \pi(x_0,x_1) = 1-(1-\eta)^{5d(2m+1)} \eqsp.
\end{align*}
Using the Bernoulli's inequality $(1+x)^{r} \geq 1+rx$ for $x \geq -1 $ and $r \geq 1$, we get
\begin{align*}
	\|\mu_{1-\eta} - \mu_1\|_{\TV} \leq 1-(1-5d(2m+1)\eta) = 5d(2m+1)\eta \lesssim dm\eta \eqsp.
\end{align*}

\paragraph{Uniform Random Walk.}
Similarly as before,
\begin{align}\label{eq:cor_1_urw}
	\| \mu_{1-\eta} - \mu_1 \|_{\TV} \leq & \P \l( \iX_{1-\eta} \neq \iX_1 \r) = 1-  \P \l( \iX_{1-\eta} = \iX_1 \r) \notag \\
	&= 1-\sum_{x_0,x_1 \in \Z^d_m} \pi(x_0,x_1) \P(X_{1-\eta} =x_1| X_0 =x_0, X_1 = x_1) \eqsp.
\end{align}
Thanks to the rigorous formula of the transition density provided in \eqref{eq:transition_density_urw}, we have for $x_0,x_1 \in \Z^d_m$:
\begin{align*}
	&\qquad \P(X_{1-\eta} =x_1| X_0 =x_0, X_1 = x_1) \\
	& =  \frac{p_{1-\eta|0}(x_1|x_0) p_{1|1-\eta} (x_1|x_1)}{p_{1|0}(x_1|x_0)} \\
	&=  \l(\frac{1-\rme^{-(1-\eta)}}{1-\rme^{-1}} \r)^{d_{\mrh}(x_0,x_1)} \l(\frac{1+(m-1) \rme^{-\eta}}{m} \r)^d \l(\frac{1+(m-1)\rme^{-(1-\eta)}}{1+(m-1)\rme^{-1}} \r)^{d- d_{\mrh}(x_0,x_1)} \eqsp.
\end{align*}
Since ${1+(m-1)\rme^{-(1-\eta)}} \geq {1+(m-1)\rme^{-1}} $ and $d-d_{\mrh}(x_0,x_1) \geq 0$, the last term above larger than 1 and we deduce that
\begin{align*}
	\P(X_{1-\eta} =x_1| X_0 =x_0, X_1 = x_1) \geq  \l(\frac{1-\rme^{-(1-\eta)}}{1-\rme^{-1}} \r)^{d_{\mrh}(x_0,x_1)} \l(\frac{1+(m-1) \rme^{-\eta}}{m} \r)^d \eqsp.
\end{align*}
Now note that $\frac{1-\rme^{-(1-\eta)}}{1-\rme^{-1}} \leq 1$ and $d_{\mrh}(x_0,x_1) \leq d$ and apply elementary inequality $x^d \geq 1-d(1-x)$ for $x \in [0,1]$, we obtain
\begin{align*}
	\P(X_{1-\eta} =x_1| X_0 =x_0, X_1 = x_1) &\geq \l( \frac{(1-\rme^{-(1-\eta)})(1+(m-1) \rme^{-\eta})}{(1-\rme^{-1})m}\r)^d \\
	&\geq 1- d \l(1 - \frac{(1-\rme^{-(1-\eta)})(1+(m-1) \rme^{-\eta})}{(1-\rme^{-1})m} \r) \\
	&= 1- \frac{d (1-\rme^{-\eta})(m-1+\rme^{-1+\eta})}{(1-\rme^{-1})m} \\
	&\geq 1 - \frac{d(1-\rme^{-\eta})}{1-\rme^{-1}}
	\geq 1- \frac{d\eta}{1-\rme^{-1}} \eqsp,
\end{align*}
where we used inequality $\rme^{-\eta} \geq 1-\eta$ in the last estimate.
Substituting it into \eqref{eq:cor_1_urw} gives
\begin{align*}
	\| \mu_{1-\eta} - \mu_1 \|_{\TV} \leq 1- \l(1- \frac{d\eta}{1-\rme^{-1}} \r) \sum_{x_0,x_1 \in \Z^d_m} \pi(x_0,x_1) = \frac{d\eta}{1-\rme^{-1}} \lesssim d\eta \eqsp.
\end{align*}
To conclude, we plug the early-stopping error in each case into
\eqref{eq:cor_main}, yielding
\begin{align*}
	\| \mathrm{Law}(\Xthetastar_{1-\eta}) - \mu_1 \|_{\TV} \lesssim \mrc(m)d\eta+ m\eta^{-1}\sqrt{\sqrt{|\Mc|} \tilde{\varepsilon} + \tilde{\varepsilon}^2 }  + \lambda(m)d m\sqrt{h \log \l(m\eta^{-1} \r) \log(\eta^{-1})},
\end{align*}
where the coefficient $\mrc(m)$ taking value $m$ for NNRW and $1$ for URW.
We complete the proof of \Cref{cor:tv_distance_2}.

\newpage

\end{document}

%% file: def_neurips.tex
\def\iX{X^{\mathrm{I}}}
\def\mX{X^{\mathrm{M}}}

\def\eqsp{\,}
\def\eqdef{:=}
\def\l{\left}
\def\r{\right}

\def\TV{\mathrm{TV}}

\def\uM{u^{\mathrm{M}}}
\def\uMtheta{u^{\mathrm{M},\theta}}
\def\uMthetastar{u^{\mathrm{M},\theta^\star}}

\def\tuMthetastar{\tilde{u}^{\mathrm{M},{\theta^\star}}}

\def\Z{\mathbb{Z}}
\def\pI{p^{\mathrm{I}}}

\def\1{\mathbf{1}}

\def\Mc{\mathcal{M}}
\def\R{\mathbb{R}}
\def\E{\mathbb{E}}
\def\Fc{\mathcal{F}}
\def\rmd{\mathrm{d}}

\def\KL{\mathrm{KL}}

\def\loss{\mathcal{L}}

\def\rme{\mathrm{e}}
\def\P{\mathbb{P}}

\def\Xthetastar{X^{\theta^\star}}

\def\N{\mathbb{N}}

\newtheorem{theorem}{Theorem}

\newtheorem{lemma}{Lemma}
\newtheorem{corollary}{Corollary}

\crefname{theorem}{theorem}{theorems}
\Crefname{theorem}{Theorem}{Theorems}

\crefname{corollary}{corollary}{corollaries}
\Crefname{corollary}{Corollary}{Corollaries}

\theoremstyle{definition}
\newtheorem{remark}{Remark}

\def\mce{\mathcal{E}}
\def\mse{\mathsf{E}}
\def\msa{\mathsf{A}}
\def\msb{\mathsf{B}}
\def\mcp{\mathcal{P}}
\def\mcam{\mathcal{A}^{\mathrm{M}}}
\def\mcbm{\mathcal{B}^{\mathrm{M}}}
\def\mccm{\mathcal{C}^{\mathrm{M}}}

\newtheorem{assumption}{\textbf{H}\hspace{-2pt}}
\Crefname{assumption}{\textbf{H}\hspace{-2pt}}{\textbf{H}\hspace{-2pt}}
\crefname{assumption}{\textbf{H}}{\textbf{H}}

\newtheorem{assumption_main}{\textbf{M}\hspace{-2pt}}
\Crefname{assumption_main}{\textbf{M}\hspace{-2pt}}{\textbf{M}\hspace{-2pt}}
\crefname{assumption_main}{\textbf{M}}{\textbf{M}}

\def\msx{\mathsf{X}}
\def\mcf{\mathcal{F}}
\def\ie{\text{i.e.}}

\def\B{\mathbb{B}}
\def\mq{q^{\mathrm{M}}}
\def\mqtheta{q^{\mathrm{M}, \theta}}
\def\mqthetastar{q^{\mathrm{M}, \theta^\star}}
\def\mrh{\mathrm{H}}
\def\mrc{\mathrm{C}}
\def\Q{\mathbb{Q}}

\def\zset{\mathbb{Z}}
\def\PP{\mathbb{P}}

\def\rset{\mathbb{R}}
\def\Q{\mathbb{Q}}
\def\W{\mathbb{W}}
\def\msi{\mathsf{I}}
\def\mre{\mathrm{e}}
\def\Mcn{\mathcal{M}^{\mathrm{N}}}
\def\Mcu{\mathcal{M}^{\mathrm{U}}}

\def\Xalg{X^{\mathrm{Alg}}}
\def\qalg{q^{\mathrm{Alg}}}